%% file: arXiv.tex
\documentclass[10pt,twocolumn,letterpaper]{article}

\usepackage[pagenumbers]{cvpr} %

\usepackage{graphicx}
\usepackage{amsmath}
\usepackage{amssymb}
\usepackage{booktabs}
\usepackage{algorithm, algorithmic}

\usepackage{amsfonts}       %
\usepackage{nicefrac}       %
\usepackage{microtype}      %
\usepackage{xcolor}         %
\usepackage{amsbsy}
\usepackage{tabularx}
\usepackage{amsmath}
\usepackage{makecell}
\usepackage{graphicx}
\usepackage{multirow}
\usepackage{dsfont}
\usepackage{wrapfig}
\usepackage{mathtools}

\usepackage{etoolbox}
\usepackage{marvosym}
\usepackage{booktabs}

\usepackage{colortbl}
\definecolor{F7E0D5}{RGB}{247,224,213}
\colorlet{Light}{white!0!F7E0D5}

\definecolor{NavyBlue}{rgb}{0.1, 0.4, 0.8}
\definecolor{Orange}{rgb}{0.9,0.5,0}
\definecolor{NavyBlue}{rgb}{0.1, 0.4, 0.8}
\definecolor{Magenta}{rgb}{0.8, 0.1, 0.6}

\newcommand{\comment}[1]{}
\usepackage{enumitem}

\newtheorem{definition}{Definition}

\newcommand{\posimprov}[1]{\scriptsize{(\textcolor{teal}{+#1})}}
\newcommand{\negimprov}[1]{\scriptsize{(\textcolor{brown}{-#1})}}

\usepackage{rotating}
\usepackage{soul}
\sethlcolor{yellow}
\setstcolor{green}
\setulcolor{red}
\usepackage[pagebackref,breaklinks,colorlinks]{hyperref}

\usepackage[capitalize]{cleveref}
\crefname{section}{Sec.}{Secs.}
\Crefname{section}{Section}{Sections}
\Crefname{table}{Table}{Tables}
\crefname{table}{Tab.}{Tabs.}

\def\cvprPaperID{1672} %
\def\confName{CVPR}
\def\confYear{2023}

\begin{document}

\title{Masked Jigsaw Puzzle: A Versatile Position Embedding for Vision Transformers}

\newcommand*\samethanks[1][\value{footnote}]{\footnotemark[#1]}

\author{Bin Ren$^{1,2}$\thanks{Equal contribution.} \ \ \  Yahui Liu$^{2}$\samethanks \ \ \  Yue Song$^{2}$ \ \ \  Wei Bi$^{3}$ \ \ \  Rita Cucchiara$^{4}$ \ \ \  Nicu Sebe$^{2}$ \ \ \  Wei Wang$^{5}$\thanks{Corresponding author.} \\
$^1$University of Pisa, Italy \quad
$^2$University of Trento, Italy\\
$^3$Tencent AI Lab , China\quad
$^5$Beijing Jiaotong University, China \\
$^4$University of Modena and Reggio Emilia, Italy
}

\maketitle

\input{latex/sections/0abstract}

\input{latex/sections/1introduction}

\input{latex/sections/2relatedworks}

\input{latex/sections/3preliminaries}
\input{latex/sections/4method}
\input{latex/sections/5experiments}
\input{latex/sections/6conclusion}

\noindent \textbf{Acknowledgments.} This work was partly supported by the National AI Ph.D. for Society Program of Italy, and the EU H2020 project AI4Media (No. 951911).

{\small
\bibliographystyle{ieee_fullname}
\bibliography{egbib}
}

\input{latex/sections/7appendix}

\end{document}

%% file: latex/sections/0abstract.tex
\begin{abstract}
    Position Embeddings (PEs), an arguably indispensable component in Vision Transformers (ViTs), have been shown to improve the performance of ViTs on many vision tasks. However, PEs have a potentially high risk of privacy leakage since the spatial information of the input patches is exposed. This caveat naturally raises a series of interesting questions about the impact of PEs on accuracy, privacy, prediction consistency, \etc. To tackle these issues, we propose a Masked Jigsaw Puzzle (MJP) position embedding method. In particular, MJP first shuffles the selected patches via our block-wise random jigsaw puzzle shuffle algorithm, and their corresponding PEs are occluded. Meanwhile, for the non-occluded patches, the PEs remain the original ones but their spatial relation is strengthened via our dense absolute localization regressor. The experimental results reveal that 1) PEs explicitly encode the 2D spatial relationship and lead to severe privacy leakage problems under gradient inversion attack; 2) Training ViTs with the naively shuffled patches can alleviate the problem, but it harms the accuracy;
    3) Under a certain shuffle ratio, the proposed MJP not only boosts the performance and robustness on large-scale datasets (\emph{i.e.}, ImageNet-1K and ImageNet-C, -A/O) but also improves the privacy preservation ability under typical gradient attacks by a large margin. The source code and trained models are available at~\url{https://github.com/yhlleo/MJP}. 
\end{abstract}
\vspace{-2em}

%% file: latex/sections/1introduction.tex
\section{Introduction}
\label{sec:introduction}

Transformers~\cite{vaswani2017attention} demonstrated their overwhelming power on a broad range of language tasks (\emph{e.g.}, text classification, machine translation, or question answering~\cite{vaswani2017attention,khan2022transformers}), and the vision community
follows it closely and extends it for vison tasks, such as image classification~\cite{dosovitskiy2020image,touvron2021training}, object detection~\cite{carion2020end,zhu2021deformable}, segmentation~\cite{ye2019cross}, and image generation~\cite{chen2021pre,liang2021swinir}. Most of the previous ViT-based methods focus on designing different pre-training objectives ~\cite{fang2021you,gidaris2018unsupervised,girdhar2019video} or variants of self-attention mechanisms~\cite{liu2021swin,wang2022pvt,wang2021pyramid}. By contrast, PEs receive less attention from the research community and have not been well studied yet. In fact, apart from the attention mechanism, how to embed the position information into the self-attention mechanism is also one indispensable research topic in Transformer. It has been demonstrated that without the PEs, the pure language Transformer encoders (\emph{e.g.}, BERT~\cite{devlin2018bert} and RoBERTa~\cite{liu2019roberta}) may not well capture the meaning of positions~\cite{wang2020position}. As a consequence, the meaning of a sentence can not be well represented~\cite{dufter2022position}. 
Similar phenomenon of PEs could also be observed in vision community. Dosovitskiy \emph{et al.}~\cite{dosovitskiy2020image} reveals that removing PEs causes performance degradation.
Moreover, Lu \emph{et al.}~\cite{lu2022april} analyzed this issue from the perspective of user privacy and demonstrated that the PEs place the model at severe privacy risk since it leaks the clues of reconstructing sequential patches back to images. Hence, it is very interesting and necessary to understand how the PEs affect the accuracy, privacy, and consistency in vision tasks. Here the consistency means whether the predictions of the transformed/shuffled image are consistent with the ones of the original image.
\begin{figure}[!t]
    \centering
    \includegraphics[width=1.0\linewidth]{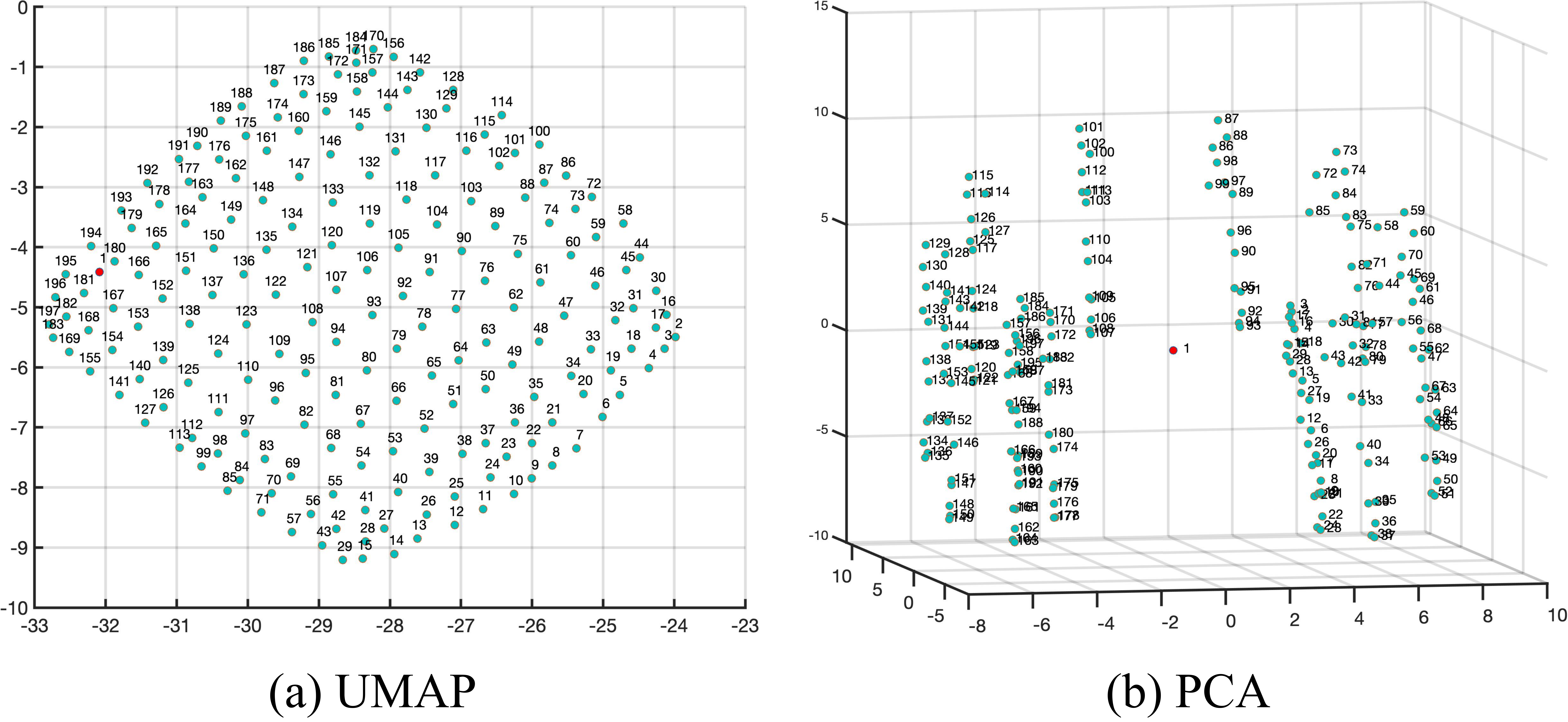}
    \vspace{-6mm}
    \caption{Low-dimensional projection of position embeddings from DeiT-S~\cite{touvron2021training}. (a) The 2D UMAP projection, it shows that reverse diagonal indices have the same order as the input patch positions. (b) The 3D PCA projection, it also shows that the position information is well captured with PEs. Note that the embedding of index $1$ (\emph{highlighted in red}) corresponds to the [CLS] embedding that does not embed any positional information.}
    \label{fig:deit_posemb}
    \vspace{-6mm}
\end{figure}

\begin{figure*}[!t]
    \centering
    \includegraphics[width=0.97\linewidth]{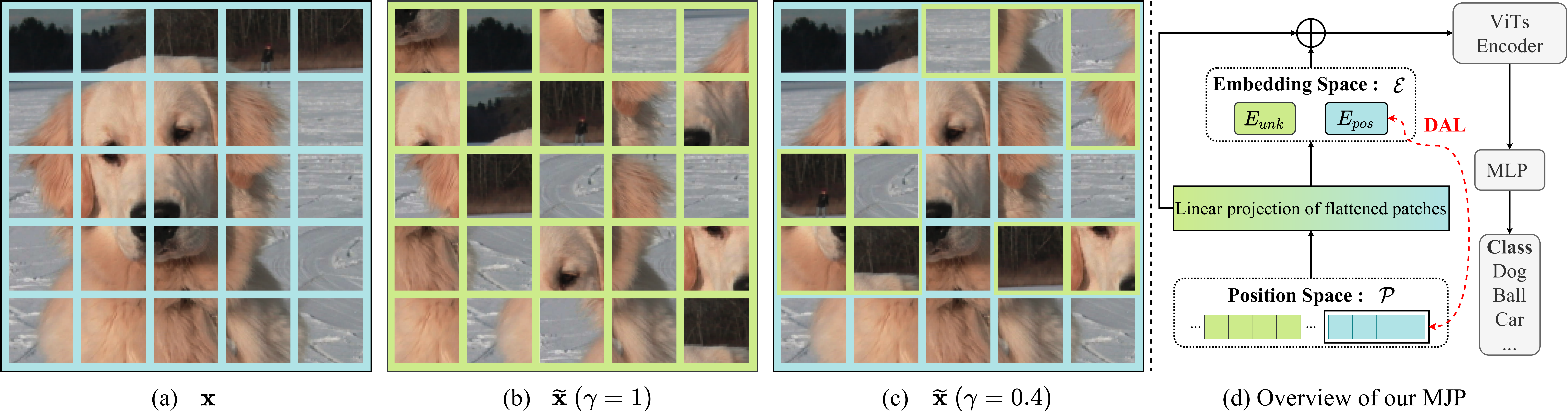}
    \vspace{-2mm}
    \caption{(a) The original input patches; (b) Totally random shuffled input patches; (c) Partially random shuffled input patches; (d) An overview of the proposed MJP. Note that we show the random shuffled patches and its corresponding \textit{unknow} position embedding in green and the rest part in blue. DAL means the self-supervised \textit{dense absolute localization} regression constraint.}
    \label{fig:overview}
    \vspace{-4mm}
\end{figure*}

To study the aforementioned effects of PEs, the key is to figure out what explicitly PEs learn about positions from input patches. To answer this question, we project the high-dimensional PEs into the 2D and 3D spaces using Uniform Manifold Approximation \& Projection (UMAP)~\cite{mcinnes2018umap} and PCA, respectively. Then for the first time, we visually demonstrate that the PEs can learn the 2D spatial relationship very well from the input image patches (the relation is visualized in Fig.~\ref{fig:deit_posemb}). We can see that the PEs are distributed in the same order as the input patch positions. Therefore, we can easily obtain the actual spatial position of the input patches by analyzing the PEs. Now it explains why PEs can bring the performance gain for ViTs~\cite{dosovitskiy2020image}. This is because the spatial relation in ViTs works similar as the inherent intrinsic inductive bias in CNNs (\emph{i.e.}, it models the local visual structure)~\cite{xu2021vitae}.
However, these correctly learned spatial relations are unfortunately the exact key factor resulting in the privacy leakage~\cite{lu2022april}.

Based on these observations, one straightforward idea to protect the user privacy is to provide ViTs with the randomly transformed (\emph{i.e.}, shuffled) input data. The underlying intuition is that the original correct spatial relation within input patches will be violated via such a transformation. Therefore, we transform the previous visually recognizable input image patches $\mathbf{x}$ shown in Fig.~\ref{fig:overview}(a) to its unrecognizable counterpart $\widetilde{\mathbf{x}}$ depicted in Fig.~\ref{fig:overview}(b) during training. The experimental results show that such a strategy can effectively alleviate the privacy leakage problem. This is reasonable since the reconstruction of the original input data during the attack is misled by the incorrect spatial relation. However, the side-effect is that this leads to a severe accuracy drop.

Meanwhile, we noticed that such a naive transformation strategy actually boosts the \textbf{consistency}~\cite{sohn2020fixmatch,xie2020unsupervised,rangrej2022consistency} albeit the accuracy drops. Note that here the consistency represents if the predictions of the original and transformed (\emph{i.e.}, shuffled) images are consistent.
Given the original input patches $\mathbf{x}$ and its corresponding transformed (\emph{i.e.}, shuffled) counterpart, we say that the predictions are consistent if $\arg\max P(\mathcal{F}(\mathbf{x}))=\arg\max P(\mathcal{F}(\widetilde{\mathbf{x}}))$, where $\mathcal{F}$ refers to the ViT models, and $P$ denotes the predicted logits. %

These observations hint that there might be a trade-off solution that makes ViTs take the best from both worlds (\emph{i.e.}, both the accuracy and the consistency). Hence, we propose the Masked Jigsaw Puzzle (MJP) position embedding method. Specifically, there are four core procedures in the MJP: (1) We first utilize a block-wise masking method~\cite{bao2021beit} to randomly select a partial of the input sequential patches; (2) Next, we apply jigsaw puzzle to the selected patches (\emph{i.e.}, shuffle the orders); (3) After that, we use a shared \textit{unknown} position embedding for the shuffled patches instead of using their original PEs; (4) To well maintain the position prior of the unshuffled patches, we introduce a \emph{dense absolute localization} (DAL) regressor to strengthen their spatial relationship in a self-supervised manner. We simply demonstrate the idea of the first two procedures in Fig.~\ref{fig:overview}(c), and an overview of the proposed MJP method is available in Fig.~\ref{fig:overview}(d). In summary, our main contributions are:
\begin{itemize}[leftmargin=*]
\vspace{-0,2cm}
    \item We demonstrate that although PEs can boost the accuracy, the consistency against image patch shuffling is harmed. Therefore, we argue that studying PEs is a valuable research topic for the community.\vspace{-0,2cm}
    \item We propose a simple yet efficient Masked Jigsaw Puzzle (MJP) position embedding method which is able to find a balance among accuracy, privacy, and consistency.\vspace{-0,2cm}
    \item Extensive experimental results show that MJP boosts the accuracy on regular large-scale datasets (\emph{e.g.}, ImageNet-1K~\cite{russakovsky2015imagenet}) and the robustness largely on ImageNet-C~\cite{hendrycks2019benchmarking}, -A/O~\cite{hendrycks2021nae}. One additional bonus of MJP is that it can improve the privacy preservation ability under typical gradient attacks by a large margin.
\end{itemize}

%% file: latex/sections/2relatedworks.tex
\section{Related Work}
\label{sec:related-work}

\subsection{Vision Transformers}
Transformers~\cite{vaswani2017attention}, originally designed for Nature Language Processing (NLP) tasks, have recently shown promising performance on computer vision tasks~\cite{han2020survey,khan2022transformers}. Benefiting from the strong representation power of modelling global relations between image patches, Vision Transformers (ViTs)~\cite{dosovitskiy2020image} have achieved superior performance than their counterpart CNNs on image classification and various other downstream tasks (\emph{e.g.}, object detection~\cite{carion2020end,zhu2021deformable}, object re-identification~\cite{he2021transreid}, dense prediction~\cite{yang2021transformer,zheng2021rethinking,wang2021end,wang2022tokencut} and image generation~\cite{chen2020generative,chen2021pre,liang2021swinir,jiang2021transgan}). 

As a core module in ViTs, multi-head self-attentions (MSAs)~\cite{vaswani2017attention,dosovitskiy2020image} aggregate sequential tokens with normalized attentions as:
$\pmb{z}_j = \sum_i \texttt{Softmax}(\frac{\pmb{Q}\pmb{K}}{\sqrt{d}})_i\pmb{V}_{i,j}$
where $\pmb{Q}$, $\pmb{K}$ and $\pmb{V}$ are query, key and value matrices, respectively. $d$ is the dimension of query and key, and $\pmb{z}_j$ is the $j$-th output token. In theory, when the position information is not considered, the outputs of MSAs should be strictly invariant to the input sequence order (\emph{i.e.}, \emph{position-insensitivity}). This indicates that a visually recognizable image can be transformed into its unrecognizable counterpart by permuting the order of image patches while maintaining the performance delivered by ViTs compared with the ViTs trained on the original non-permuted image. However, the usage of PEs hinders such implementations, where the outputs of the ViTs vary dramatically with the mentioned naive transformations shown in Fig.~\ref{fig:overview}(b). In this work, our main focus is to explicitly figure out what PEs actually learn from input patches about positions, and how the PEs affect the accuracy, privacy, and the consistency properties of ViTs.

\subsection{Position Embeddings}
In Transformer networks, both the attention and the (individual token based) feed-forward layers are permutation invariant when the position information is not considered. In this way, the spatial relationships between image patches could not be modeled as the position information is completely discarded. As compensation, PEs are naturally introduced into ViTs to provide information about the token order during the learning process, since it offers possibilities for dependency modeling between elements at different positions. For example, previous works~\cite{gehring2017convolutional,vaswani2017attention,shaw2018self} indicated that PEs are useful to give the model a sense of which portion of the sequence in the input/output it is currently dealing with. Inspired by this, some works~\cite{jung2020global,su2021roformer,kiyono2021shape,liu2022petr,he2022masked} showed diverse application scenarios that benefit from the usage of suitable PEs. In addition, Chu \emph{et al.}~\cite{chu2021conditional} proposed correlating the PEs with their local neighborhood of the input sequence. Liu \emph{et al.}~\cite{liu2021efficient} proposed to enhance the spatial prior (\emph{i.e.}, relative localization) in the final content embedding to indirectly enrich the inductive bias. Obviously, although these methods enhance the position information learnt by PEs, they indeed degenerate the position-insensitive property of MSAs.

Especially, Wang \emph{et al.}~\cite{wang2020position} revealed that Transformer encoders (\emph{e.g.}, BERT~\cite{devlin2018bert} and RoBERTa~\cite{liu2019roberta}) may not well capture the meaning of positions (absolute and relative positions). They showed that Transformer encoders learn the local position information that can only be effective in masked language modeling. In contrast, there does not exist such a similar "masked language modeling" procedure in ViTs (and VTs) in the typically supervised pre-training. Moreover, Lu \emph{et al.}~\cite{lu2022april} revealed that the learnable PEs place the model at severe privacy risk, which leaks the clues of reconstructing sequential patches to images. In this paper, we dive into the usage of PEs and propose an efficient position embedding method, MJP, to improve the position-insensitive property of ViTs without hurting the positive effects of PEs.

%% file: latex/sections/3preliminaries.tex
\section{Preliminaries: 2D/3D Spatial Priors in PEs}
\label{sec:preliminaries}

Although numerous works claim that the PEs can learn the 2-D spatial relationship of image patches, this claim has not been demonstrated visually or mathematically. To visualize the concrete relation of image patches captured in the high-dimensional position embedding, we project them into the 2-D and 3-D spaces using Uniform Manifold Approximation and Projection (UMAP)~\cite{mcinnes2018umap} and PCA, respectively. Fig.~\ref{fig:deit_posemb} displays the projections of PEs from DeiT-S~\cite{touvron2021training}. Note that~\cite{dosovitskiy2020image} only shows the cosine similarity between the PEs, which is not exactly the 2D spatial information mentioned in our paper. Here, we explicitly specify the spatial relationship as the relative distance in the 2D/3D coordinate space. As shown in Fig.~\ref{fig:deit_posemb}, the spatial relationship can be projected into a 2D/3D coordinate system and indicate the localizations of these embeddings. For the UMAP, the projected positional embedding emerges as grid-like structures with the distances between each point roughly the same, which is coherent 
with the relation of embedded image patches. For the PCA, the position embedding that corresponds to neighboring image patches groups together and aggregates to the stick-like form. This phenomenon demonstrates that the spatial relationship is indeed learned by the PEs. Moreover, this also implies that \emph{the spatial relationship learned in the high-dimensional space still manifests in the low-dimensional space}.

The learnable PEs, which work as a lookup table of a dictionary, maps the 1-dimensional data into the sparse high-dimensional space. By nature, it is a sparse large matrix. Dimensionality reduction techniques can capture the structural information of sparse matrices using low dimensionalities. For the PCA, the amount of information retained in the projection can be measured by the ratio of explained variance. Formally, we have the definition as follows:

\vspace{-2mm}
\begin{definition}[Explained Variance]
\label{def:explain_variance}
    Let $\mathbf{P}$ and $\mathbf{V}$ denote the data matrix and the PCA projection matrix. The ratio of explained variance that $\mathbf{PV}$ accounts for is defined as $\nicefrac{\sum\sigma(\mathbf{PV})^2}{\sum\sigma(\mathbf{P})^2}$ where $\sigma(\cdot)$ denotes the singular value.
\end{definition}
\vspace{-2mm}

In practice, we observe that the 3-dimensional PCA projection explains $54.6\%$ of the total variance of the DeiT-S embedding matrix. Given that a large amount of information can be captured by the low-dimensional projection, we propose that explicitly enforcing low-dimensional positional prior can help the positional learning in the high-dimensional space, which might accelerate the convergence rate of training and improve the performance (See Sec. ~\ref{subsec:dal}).

%% file: latex/sections/4method.tex
\section{Method}
\label{sec:method}
Based on the design principle of MSAs, the outputs of MSAs should be entirely position-agnostic. However, PEs hinder such a property because it can learn the strong 2-D spatial relation of the input patches (as we demonstrated in Sec.~\ref{sec:preliminaries}). To this end, we propose the block-wise random jigsaw puzzle shuffle algorithm (See Alg.~\ref{alg:1}) to transform the input patches with different shuffle ratios $\gamma$ for intermingling the original correct spatial relation. 

\begin{algorithm}[t]
	\renewcommand{\algorithmicrequire}{\textbf{Input:}}
	\renewcommand{\algorithmicensure}{\textbf{Output:}}
	\caption{Block-wise Random Jigsaw Puzzle Shuffle}
	\label{alg:1}
	\begin{algorithmic}[1]
		\REQUIRE Input image: $\mathbf{x}\in \mathbb{R}^{H\times W \times C}$; \\  ~~~~~~Shuffle Ratio: $\gamma$; \\
		~~~~~~Patch Size: $P$
		\ENSURE Shuffled image patches: $\widetilde{\pmb{x}}_p$
		\STATE 
		~~~~~~$\mathbf{x}_p\in\mathbb{R}^{N\times (P^{2} \cdot C)} \longleftarrow Patchlize(\mathbf{x}, P)$
		\STATE ~~~~~~$\mathbf{m}\in\mathbb{R}^{\frac{H}{P}\times\frac{W}{P}} \longleftarrow BinaryInitialize(\mathbf{x}_p, 0)$
		\STATE 
		~~~~~~$\widetilde{\mathbf{m}} \in\mathbb{R}^{\frac{H}{P}\times\frac{W}{P}} \longleftarrow BlockwiseMask(\mathbf{m}, \mathbf{\gamma})$~\cite{liu2019roberta}
	    \STATE
	    ~~~~~~$\widetilde{\pmb{x}}_p \in\mathbb{R}^{N\times (P^{2} \cdot C)} \longleftarrow JigsawPuzzle(\mathbf{x}_p, \widetilde{\mathbf{m}})$
		\STATE 
		\textbf{return} $\widetilde{\pmb{x}}_p$
	\end{algorithmic}  
	\vspace{-0.2em}
\end{algorithm}

Since we experimentally demonstrate that the totally shuffled strategy (\emph{i.e.}, $\gamma=1.0$) will degenerate the accuracy a lot albeit the consistency increases. As a remedy, we only shuffle portion of the sequence patches, and \emph{strengthen} the spatial relation of the rest part with a dense absolute localization regression. Finally, a versatile position embedding method MJP is proposed. The detailed analysis of each module is available in the ablation study in Sec.~\ref{subsec:ablation}. In the following sub-sections, we will introduce the Jigsaw Puzzle Shuffle method (Sec.~\ref{sec:block_shuffle}), the spatial relation strengthen method (Sec.~\ref{subsec:dal}), and our final MJP method (Sec.~\ref{subsec:mjp}).

\subsection{Block-wise Random Jigsaw Puzzle Shuffle}
\label{sec:block_shuffle}
Specifically, given an input image $\mathbf{x}\in \mathbb{R}^{H\times W \times C}$, we first reshape it into a sequence of flattened 2D patches $\mathbf{x}_p\in\mathbb{R}^{N\times (P^{2} \cdot C)}$, where $(H, W)$ is the resolution of the original image, $C$ is the number of channels, $(P, P)$ is the resolution of each image patch, and then we have $N = HW/P^2$, which denotes the number of patches. Then instead of directly applying block-wise masking method~\cite{bao2021beit} to the image, we first initialize a binary mask matrix $\mathbf{m}\in\mathbb{R}^{\frac{H}{P}\times\frac{W}{P}}$ with the same size as the image patches in $\mathbf{x}_p$. Next, we use~\cite{bao2021beit} to update the binary mask $\mathbf{m}$ in which the masked positions will be set to 1 and the rest untouched positions remains 0. The hyper-parameter $\gamma$ is used to control the ratio of selected positions. After that, a jigsaw puzzle shuffle operation is applied to $\mathbf{x}_p$ conditioned on the updated binary mask $\widetilde{\mathbf{m}}$. 

Finally, we get the shuffled patch sequence $\widetilde{\mathbf{x}}_p$, where the shuffled patches are actually $\{\mathbf{x}_p^i | \widetilde{\mathbf{m}}_i = 1\}$. Notably, in~\cite{bao2021beit}, the patches covered by the sampled mask are not visible to the encoder module, while in our method, the masking strategy~\cite{bao2021beit} is only used to mask out the positions/indices of the selected patches. These patches are still visible to the encoder module and they are randomly shuffled in a jigsaw puzzle manner. One intuitive shuffled toy example is shown in Fig.~\ref{fig:overview}(c), where the green part means the selected shuffled region while the blue part remains unchanged. Actually, this algorithm shares a similar idea with the data augmentation method as the training images are always changing as their patches are randomly shuffled during each iteration.

\subsection{Strengthening Spatial Prior in PEs}
\label{subsec:dal}
Liu \emph{et al.}~\cite{liu2021efficient} noticed that by \emph{enhancing} the 2-D spatial information of the output embeddings of the last layer of ViTs, the training convergence speed can be accelerated. Inspired by their work, we observe that similar gains can be obtained by applying a low-dimensional spatial prior in PEs. Different from the dense relative localization constraint in~\cite{liu2021efficient} which samples relative pairs from the whole output sequential embeddings, we propose a much simpler \emph{dense absolute localization} (DAL) regression method to directly use self-supervised absolute location to enhance the spatial information in PEs, which avoids sampling relative pairs.

Since the PEs capture the absolute position of the input patches, to some extent, the position information could be reconstructed via a reversed mapping function $g(\cdot):\mathcal{E} \to \mathcal{P}$, where $\mathcal{E}$ and $\mathcal{P}$ are embedding space and position space, respectively. Given that PEs have one-to-one correspondence with the sequential image patches, we can reshape them into $\mathbf{E}_{\text{pos}}\in\mathbb{R}^{K\times K \times D}$, where $K$ refers to height/weight of the grid and $D$ refers to latent vector size. Then we can compute the reverse mapping from $\mathcal{E} \to \mathcal{P}$ via:
\begin{equation} 
(\widetilde{i},\widetilde{j})^T = g(\mathbf{E}^{i,j}_{\text{pos}}),
\end{equation}
in which $(\widetilde{i},\widetilde{j})^T$ is the predicted patch position, and $\mathbf{E}^{i,j}_{\text{pos}}$ is the position embedding of the patch $(i,j)$ in the $K\times K$ grid.
The dense absolute localization (DAL) loss is:
\begin{equation}
\mathcal{L}_{\text{DAL}} = \mathbb{E}_{\mathbf{E}^{i,j}_{\text{pos}}, 1\leq i,j \leq K}[\|(i,j)^T - (\widetilde{i},\widetilde{j})^T\|_1],
\end{equation}
where the expectation is computed by averaging the $\ell_1$ loss between the correspond $(i, j)^T$ and $(\widetilde{i},\widetilde{j})^T$. Then,  $\mathcal{L}_{\text{DAL}}$ is added to the standard cross-entropy loss ($\mathcal{L}_{\text{CE}}$) of the native ViTs. The final loss is: $\mathcal{L}_{\text{all}} = \mathcal{L}_{\text{CE}} + \lambda \mathcal{L}_{\text{DAL}}$, where we set $\lambda =$ 0.01 for all experiments. Note that the mapping function can be either linear or nonlinear. Throughout this work, we mainly discuss three implementations, including non-parametric PCA, learnable linear (LN), and nonlinear (NLN) projection layers. They will be discussed in details in Sec.~\ref{subsec:ablation}.

\subsection{MJP Position Embedding}
\label{subsec:mjp}
The main goal of MJP position embedding in our work is to enhance the consistency (\emph{i.e.}, position-insensitive property) of ViTs and preserve the accuracy (keeping the spatial relation modeling functionality of PEs) for the standard classification tasks.
Usually, with original input image patches $\mathbf{x}$ we can formulate the function of the input layer which is located before the Transformer block as follows:
\begin{equation}
\mathbf{z}_{0} = [\mathbf{x}_{\text{CLS}}; \mathbf{x}_p^1\mathbf{E}; \mathbf{x}_p^2\mathbf{E}, \cdots, \mathbf{x}_p^N\mathbf{E}] + \mathbf{E}_{\text{pos}},
\end{equation}
where $\mathbf{E}\in\mathbb{R}^{(P^2\cdot C)\times D}$ is a trainable linear projection layer
and $\mathbf{x}_p^k\mathbf{E}$ refer to the output of linear projection (\ie, the patch embeddings), 
and $\mathbf{E}_{\text{pos}}\in\mathbb{R}^{(N+1)\times D}$ refers to PEs (the additional one is applied to the [CLS] embedding). 

Next, we apply the proposed block-wise random jigsaw puzzle shuffle algorithm to $\mathbf{x}_p$ and produce the transformed patch sequences $\widetilde{\mathbf{x}}_p$ with the shuffle ratio $\gamma$. In this scenario, if we maintain the original position embedding sequence, it leads to a mismatch issue between the shuffled patches and the position embedding sequence. Therefore, we introduce a shared \emph{unknown} position embedding to the shuffled positions to alleviate the mismatching issue. With the corresponding updated mask $\widetilde{\mathbf{m}}$ from Alg.~\ref{alg:1} (1st \& 2nd procedures), we propose our MJP PEs:
\begin{equation}
\widetilde{\mathbf{E}}_{\mathrm{pos}}^i= \begin{cases}\mathbf{E}_{\mathrm{pos}}^i, & \text { if } \widetilde{\mathbf{m}}_i=0 \\ \mathbf{E}_{\mathrm{unk}}, & \text { if } \widetilde{\mathbf{m}}_i=1\end{cases}
\end{equation}
where $\mathbf{E}_{\text{unk}}\in\mathbb{R}^{1\times D}$ refers to a share learnable embedding (\emph{i.e.}, \emph{unknown} position embedding). It represents that the image patch in this position has random permutation and its position should be occluded (3rd procedure). $\mathbf{E}_{\text{pos}}$ is the original position embedding for the rest image patches.

\begin{algorithm}[!t]
	\renewcommand{\algorithmicrequire}{\textbf{Input:}}
	\renewcommand{\algorithmicensure}{\textbf{Output:}}
	\caption{The pipeline of the proposed MJP.}
	\label{alg:2}
	\begin{algorithmic}[1]
		\REQUIRE Input image: $\mathbf{x}\in \mathbb{R}^{H\times W \times C}$; \\  ~~~~~~Shuffle Ratio: $\gamma$; Patch Size: $P$ 
		\STATE 
		$\widetilde{\pmb{x}}_p \leftarrow Alg. ~\ref{alg:1}(\mathbf{x}, P, \gamma)$ // ~\small{1st \& 2nd procedures}
		\STATE 
		$\mathbf{E}_{\mathrm{unk}}(\widetilde{\pmb{x}}_p)$ // ~\small{3rd procedure}
		\STATE
		$\mathbf{DAL}(\mathbf{x} - \mathbf{x} \cap \widetilde{\pmb{x}}_p)$ // \small{~4th procedure, only for \textbf{training}}
	\end{algorithmic}  
	\vspace{-0.3em}
\end{algorithm}

Besides, we revisit the remaining PEs (\emph{i.e.}, corresponding to the non-selected patches) and apply low-dimensional prior on them. The low-dimensional prior is imposed by the proposed DAL (Sec.~\ref{subsec:dal}) regression method for strengthening the spatial relation  (4th procedure).
A toy illustration of the proposed MJP is shown in Fig.~\ref{fig:overview}(d), where the green color represents the randomly permuted patches and its corresponding \emph{unknown} PEs, while the blue color indicates the rest regular patches and its related original PEs. We also formalize these procedures as an algorithm in Alg.~\ref{alg:2}.

Thus, we replace the input layer with a new formulation:
\begin{equation}
\widetilde{\mathbf{z}}_{0} = [\mathbf{x}_{\text{CLS}}; \widetilde{\mathbf{x}}_p^1\mathbf{E}; \widetilde{\mathbf{x}}_p^2\mathbf{E}, \cdots, \widetilde{\mathbf{x}}_p^N\mathbf{E}] + \widetilde{\mathbf{E}}_{\text{pos}}.
\end{equation} 
the following procedures and modules are exactly the same as the ones in the original ViTs. 

%% file: latex/sections/5experiments.tex
\section{Experiments}
\label{sec:experiments}
We follow the typical supervised pre-training procedure, where all the compared models are trained on ImageNet-1K~\cite{russakovsky2015imagenet} to show the capacity of our proposed MJP method. We also benchmark the proposed MJP method on ImageNet-1K, which contains 1.28M training images and 50K validation images of 1,000 classes. The training details mostly follow the training protocols\footnote{\url{https://github.com/facebookresearch/deit}} from Touvron \emph{et al.}~\cite{touvron2021training}.

\subsection{Regular ImageNet-1K training}
We mainly compare with three typical existing methods, including two state-of-the-art Visual Transformers (\emph{i.e.}, DeiT~\cite{touvron2021training} and 
Swin~\cite{liu2021swin}) and one widely-used CNN-based ResNet-50~\cite{he2016deep}. All these methods are of comparable sizes (i.e., less than 30M parameters). Besides the common Top-1 accuracy (\textbf{Top-1 Acc.}), we also report another two evaluation metrics (\ie Diff. Norm. and Consistency) to show the position invariance of a model to the jigsaw puzzle transformation. For a given input image $\mathbf{x}$, we collect its counterpart $\widetilde{\mathbf{x}}$ by applying masked jigsaw puzzle to shuffle a portion of selected patches. Next, we calculate the difference $\ell_2$-norm (\emph{i.e.}, \textbf{Diff. Norm.}) between the [CLS] embedding inferred from $\mathbf{x}$ and $\widetilde{\mathbf{x}}$: 
\vspace{-2mm}
\begin{equation}
    \|\mathcal{F}_{\text{CLS}}(\mathbf{x}) - \mathcal{F}_{\text{CLS}}(\widetilde{\mathbf{x}})\|_2^2.
\end{equation}
For \textbf{Consistency}, we measure how many classification results stay the same (\%) given the perturbation: 
\begin{equation}
    \mathbb{E}_{\mathbf{x}}[\mathds{1}\{\arg\max P(\mathcal{F}(\mathbf{x}))=\arg\max P(\mathcal{F}(\widetilde{\mathbf{x}}))\}].
\end{equation} 
$P(\mathcal{F}(\mathbf{x}))$ denotes the predicted classification probability of image $\mathbf{x}$, and $\arg\max P(\mathcal{F}(\mathbf{x}))$ represents the predicted class index.
For a fair comparison, we use $\gamma=0.15$ to create $\widetilde{\mathbf{x}}$ in Table~\ref{tab:imagenet-pretraining}. According to these results, the proposed MJP does not have a \textit{negative} effect on the top-1 accuracy (MJP even brings a marginal improvement for DeiT-S). More importantly, it improves Diff. Norm. and Consistency by a large margin.
Besides, MJP works well in the variants of ViTs (\emph{e.g.}, Swin~\cite{liu2021swin}), which shows a good potential generalization ability. Note that we add MJP to ResNet50 by only shuffling the image patches with Alg.~\ref{alg:1} (No PEs). For Swin-T, we add MJP to the absolute PEs of Swin-T.

\begin{table}[t]
  \centering
  \caption{Comparisons of different backbones on ImageNet-1K classification. Note that the image size here are all set to 224x224.}
  \vspace{-1em}
  \resizebox{1.0\columnwidth}{!}{
  \setlength{\tabcolsep}{0.1em} 
  \begin{tabular}{XlccccX}
    \toprule
     \textbf{Method} & \textbf{Param.} & \textbf{Top-1 Acc.} $\uparrow$ & \textbf{Diff. Norm.} $\downarrow$ & \textbf{Consistency} $\uparrow$ \\ \midrule 
     ResNet-50~\cite{he2016deep} & 25 & 79.3 & 11.77 & 51.5 \\
     \rowcolor{Light}ResNet-50 + MJP & 25 & 79.4 & 7.11 & 69.3 \\
     DeiT-S~\cite{touvron2021training} & 22 & 79.8 & 16.21 & 64.3 \\
     \rowcolor{Light}DeiT-S + MJP & 22 & 80.5 & 8.96 & 82.9 \\
     Swin-T~\cite{liu2021swin} & 29 & 81.3 & 15.49 & 41.5 \\
     \rowcolor{Light}Swin-T + MJP & 29 & 81.3 & 12.36 & 66.9 \\
    \bottomrule
  \end{tabular}
  \label{tab:imagenet-pretraining}
  }
  \vspace{-1em}
\end{table}

\begin{table}[!t]
  \centering
  \caption{Comparisons on robustness to common corruptions and adversarial examples. }
  \vspace{-1em}
  \resizebox{1.\columnwidth}{!}{
  \setlength{\tabcolsep}{0.5em} 
  \begin{tabular}{lcccccccc}
    \toprule
     \textbf{Method} & \textbf{ImageNet-C} & \multicolumn{2}{c}{\textbf{ImageNet-A}} & \textbf{ImageNet-O} \\  \cmidrule(lr){3-4} 
     & mCE $\downarrow$ &  Acc $\uparrow$ & AURRA $\uparrow$ & AUPR $\uparrow$  \\ \midrule
     DeiT-S & 54.6 &  19.2 & 25.1 & 20.9 \\
     \rowcolor{Light}DeiT-S + MJP & \textbf{40.78} &  \textbf{21.6} & \textbf{29.8} & \textbf{22.6} \\
    \bottomrule
  \end{tabular}
  \label{tab:ablation-robustness}
  }
  \vspace{-1em}
\end{table}

\begin{figure*}[!ht]
\vspace{-1em}
    \centering
    \begin{subfigure}[b]{0.3\textwidth}
        \centering
        \includegraphics[width=0.9\linewidth]{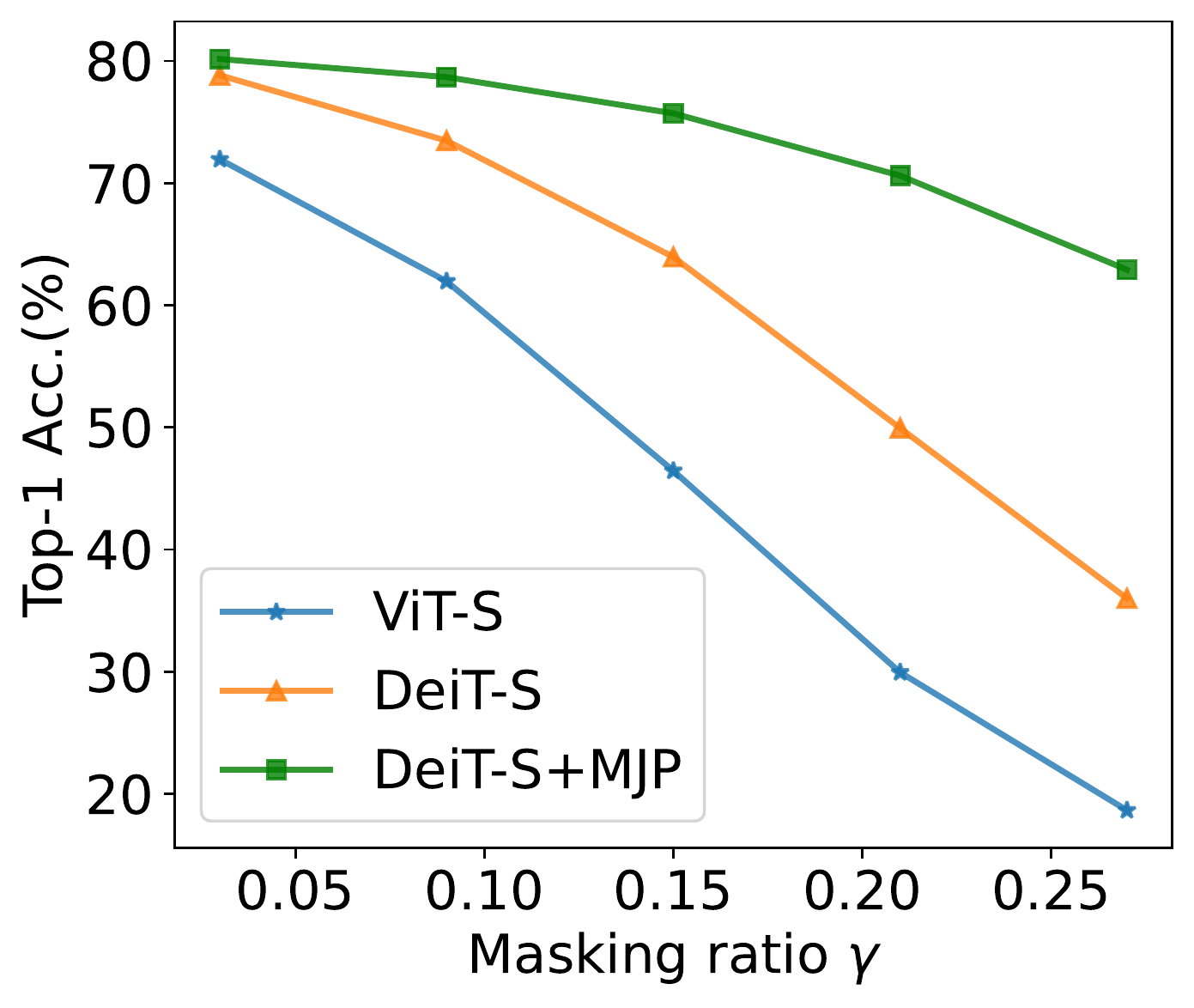}
        \caption{}
    \end{subfigure}
    \hfill
    \begin{subfigure}[b]{0.3\textwidth}
        \centering
        \includegraphics[width=0.9\linewidth]{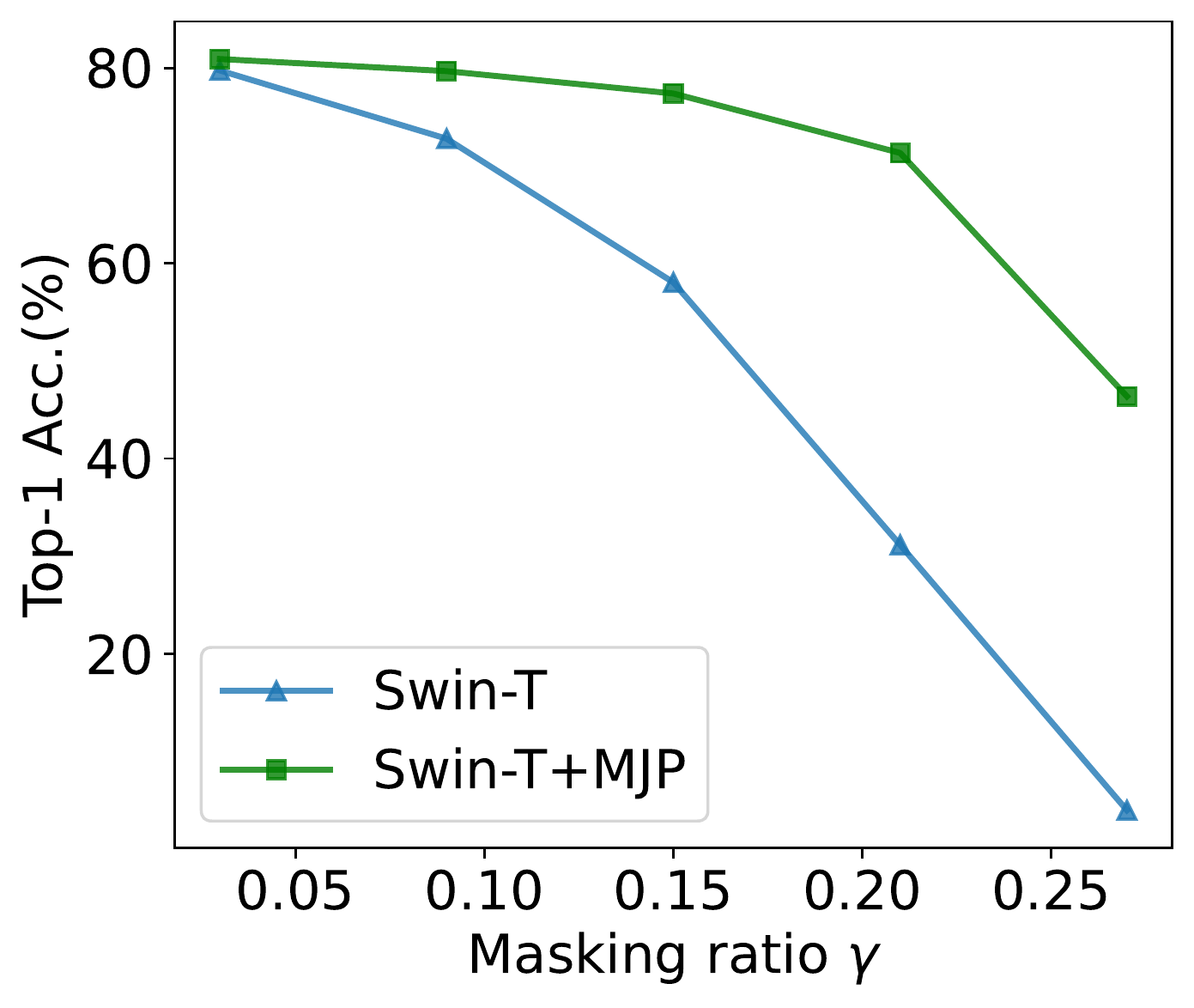}
        \caption{}
    \end{subfigure}
    \hfill
    \begin{subfigure}[b]{0.3\textwidth}
        \centering
        \includegraphics[width=0.9\linewidth]{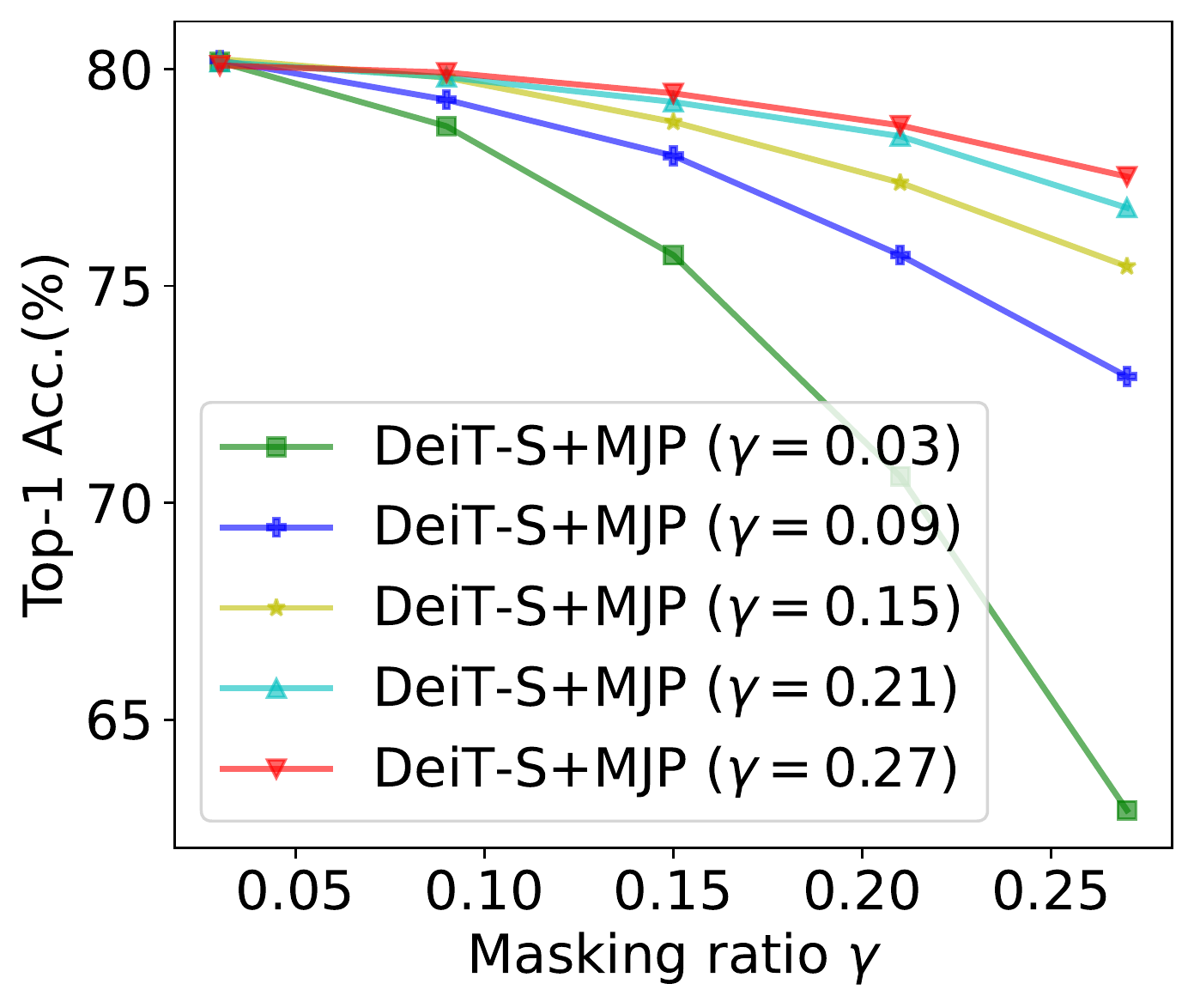}
        \caption{}
    \end{subfigure}
    \vspace{-3mm}
    \caption{Ablation on the mask ratio $\gamma$ during inference: (a) comparisons among ViT-S, DeiT-S and our method (trained with $\gamma=0.03$); (b) comparisons between Swin-T and our method (trained with $\gamma=0.03$); (c) comparisons of our method on DeiT-S trained with different $\gamma$.}
    \label{fig:masking-ratio}
    \vspace{-1.5em}
\end{figure*}

\subsection{Robustness on Challenging Sets}
Besides the strength on standard classification, we also observe an auxiliary benefit of the proposed MJP on robustness.
ImageNet-C~\cite{hendrycks2019benchmarking}\footnote{In the camera-ready version, we reported the unnormalized results. However, in accordance with the conventions of the robustness literature, we updated this version to include the normalized values. Nonetheless, it is important to note that our MJP consistently outperforms the DeiT-S model for both evaluation methods.} benchmarks a classifier’s robustness to common corruptions\footnote{\url{https://github.com/hendrycks/robustness}}. The mean Corruption Error (mCE) is used to measure the generalization of a model at corrupted distributions (the lower the better).

ImageNet-A/O~\cite{hendrycks2021nae} focus on adversarial examples, and it enables us to either test image classification performance when the input data distribution shifts (\emph{i.e.}, Acc and AURRA), or test out-of-distribution detection performance when the label distribution shifts (\emph{i.e.}, AUPR)\footnote{\url{https://github.com/hendrycks/natural-adv-examples}}. 
As shown in Table~\ref{tab:ablation-robustness}, the proposed MJP not only boosts the accuracy on the standard evaluation on ImageNet-1K validation set but also improves the robustness largely on the adversarial samples. 

The underlying reason might be that MJP enforces the ViTs aware of both local and global context features, and it helps ViTs to get rid of some unnecessary sample-specific local features during the training. This has been verified by the visualization maps (see Fig.~\ref{fig:attention}).

\subsection{Ablation Analysis}
\label{subsec:ablation}
\noindent\textbf{Results with different MJP ratios.}
As shown in Table~\ref{tab:ablation-MJP-ratio}, we test different masking ratios used in the block-wise masking strategy during the training. Obviously, Diff. Norm. has the inverse tendencies compared to Consistency, where a smaller Diff. Norm. usually indicates a larger (better) consistency score. Comparing the accuracy trained with $\gamma > 0$, it shows our model is not sensitive to different $\gamma$. In particular, a small ratio (\emph{e.g.}, $\gamma = 0.03$) is sufficient for boosting the accuracy. In addition, a large ratio can reduces the Diff. Norm. and improves the consistency by a large margin. Moreover, the model trained with a larger $\gamma$ is inclined to be more consistent as shown in Fig.~\ref{fig:masking-ratio} (c). In addition, we observe that the consistency keeps increasing when the mask ratio increases from 3\% to 27\%. However, further increasing the mask ratio will not bring consistency improvement and what is worse is that the accuracy marginally decreases.

\begin{table}[!t]
  \centering
  \caption{Ablation study on the proposed MJP method trained with different masking ratio $\gamma$.}
  \vspace{-0.8em}
  \resizebox{1.\columnwidth}{!}{
  \begin{tabular}{ccccccc}
    \toprule
     \multirow{2}{*}{\textbf{Metric}} & \multicolumn{6}{c}{\textbf{Masking Ratio}}  \\  \cmidrule(lr){2-7} 
     & 0 & 0.03 & 0.09 & 0.15 & 0.21 & 0.27 \\ \midrule
     Top-1 Acc. & 80.0 & 80.5 & 80.3 & 80.4 & 80.2 & 80.3 \\
     Diff. Norm. & 16.56 & 8.96 & 6.36 & 5.23 & 4.39 & 3.97 \\
     Consistency & 64.0 & 82.9 & 88.1 & 90.5 & 92.3 & 93.1 \\
    \bottomrule
  \end{tabular}
  \label{tab:ablation-MJP-ratio}
  }
  \vspace{-1em}
\end{table}

For a model trained with a fixed $\gamma$, we also test its accuracy with different masking ratios during the inference. As shown in Fig.~\ref{fig:masking-ratio} (a) and (b), the performances of original models drop significantly when we shuffle more image patches. In contrast, our proposed method shows more consistent performances.

\noindent\textbf{Comparison of variants of MJP.} We test several variants of the proposed MJP, including (1) Removing the PEs from the original DeiT-S; (2) SPP: shuffling both the 16$\times$16 patches and pixels within the patches, which is tested in MLP-Mixer~\cite{tolstikhin2021mlp} to validate the invariance to permutations; (3) JP: applying masked jigsaw puzzle to the sequence patches; (4) IDX: using an additional collection of embeddings to indicate the global indexes for the input patches; (5) UNK: replacing the PEs in the masked positions with a shared \emph{unknown position embedding}; (6) DAL: jointly learning \emph{dense absolute localization} regression in an self-supervised manner during the pretraining, where we provide PCA, linear (LN) and nonlinear (NLN) projections, respectively.  

\begin{table}[t]
  \centering
  \caption{Ablation study on the variants of the proposed MJP.}
  \vspace{-0.5em}
  \setlength\tabcolsep{10pt}
  \setlength{\tabcolsep}{0.3em} 
  \resizebox{1.\columnwidth}{!}{
  \begin{tabular}{llc}
    \toprule
     \textbf{Method} & \textbf{Top-1 Acc.} $\uparrow$ & \textbf{Consistency} $\uparrow$ \\ \midrule
     A: DeiT-S~\cite{touvron2021training} & 79.8 & 64.3 \\
     B: A - PEs & 77.5 \negimprov{2.3} & 100.0 \\
     C: A + SPP~\cite{tolstikhin2021mlp} & 74.9 \negimprov{4.9} & 74.8 \\
     D: A + DAL (NLN) & 80.0 \posimprov{0.2} & 64.0  \\
     E: A + JP & 79.2 \negimprov{0.6} &  73.8 \\
     F: A + JP + IDX & 79.9 \posimprov{0.1}  &   79.6 \\
     G: A + JP + UNK & 80.1 \posimprov{0.3}  &  83.8 \\
     H: A + JP + UNK + DAL (PCA) & 79.9 \posimprov{0.1}  & 83.4 \\
     I: A + JP + UNK + DAL (LN) & 80.0 \posimprov{0.2}  & 83.8 \\
     J: A + JP + UNK + DAL (NLN) & 80.5 \posimprov{0.7}  & 82.9 \\
    \bottomrule
  \end{tabular}
  \label{tab:ablation-MJP}
  }
\vspace{-0.5em}
\end{table}

Table~\ref{tab:ablation-MJP} shows the detailed ablation studies on the variants of our proposed MJP method. First, as expected, when we remove the PEs from the original DeiT-S model, the accuracy decreases by 2.3\%, but the consistency achieves 100\%. It verifies that the ViTs are naturally position-insensitive once without using PEs. 

We observe that the previous SPP strategy harms the accuracy of the model (\emph{i.e.}, $-$4.9\%), which indicates it is insufficient to simply shuffle both patches and pixels in the input image. As we expect, the usage of UNK embedding alleviates the confusion between shuffled and unshuffled positions, which boosts both the accuracy and consistency.

Finally, we can find that using \emph{dense absolute localization} regression on the unmasked PEs can marginally boost the accuracy. Although PCA does not involve more parameters to the model, it significantly increases the computation latency (e.g., $+$46\% with only 5 iterations). In general, it is better to use a nonlinear projection (\emph{e.g.}, a 3-layers MLP with negligible computation latency) to allow the learned PEs to aggregate more additional information.

\begin{figure}[!t]
\centering
\vspace{-0.2em}
\includegraphics[width=0.88\columnwidth]{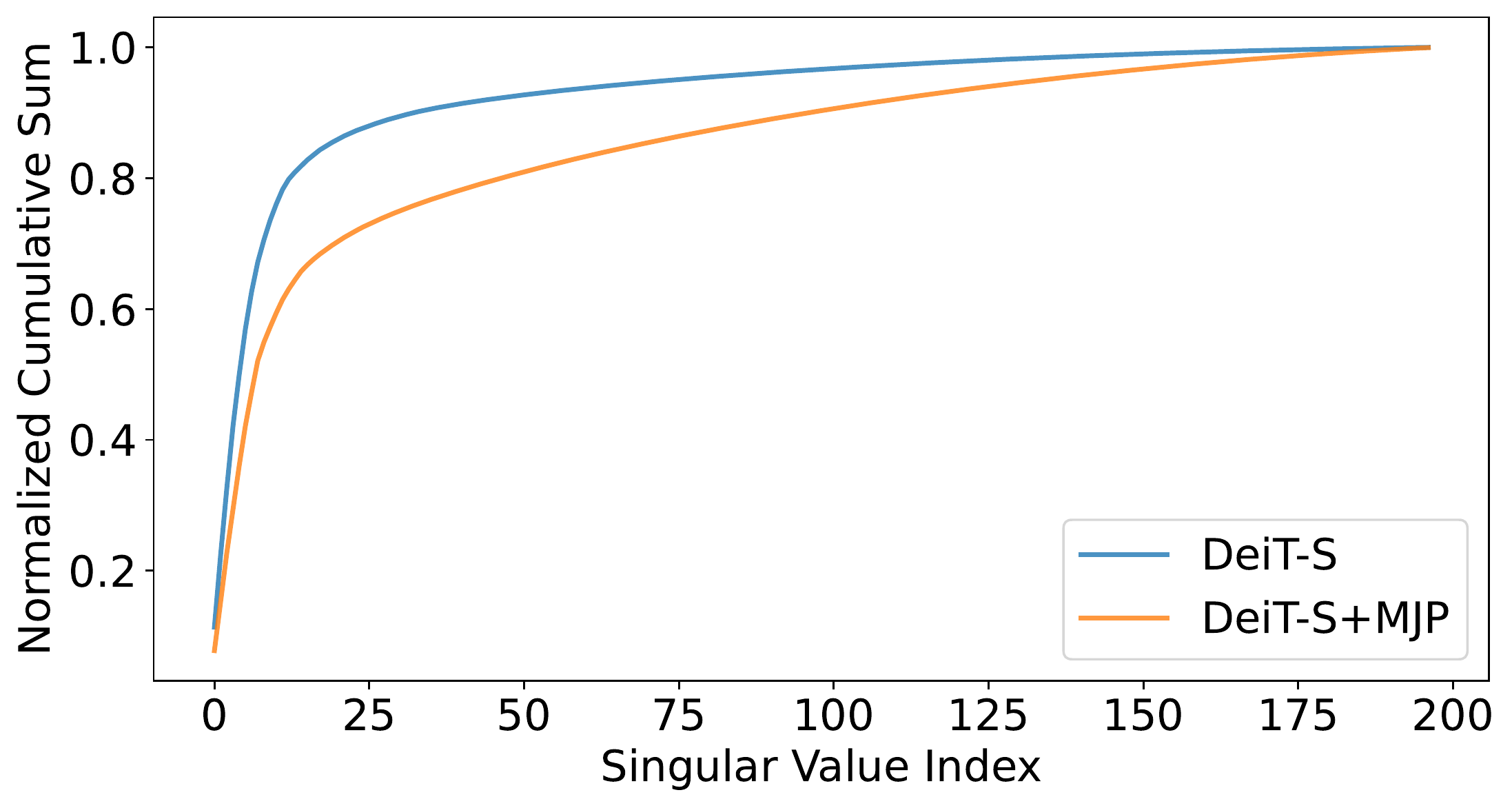}
\vspace{-1em}
\caption{Distributions of accumulated eigenvalues of PEs.}
\label{fig:eigenvalues}
\vspace{-1.5em}
\end{figure}

\vspace{2mm}
\noindent\textbf{Informativeness of the PEs.} For a position space $\mathcal{P}$ (\emph{i.e.}, a 1-dim or 2-dim space), we may not need a high-dimension embedding space $\mathcal{X}$ to model the positions. To measure the informativeness of the learned PEs, we apply singular value decomposition (SVD) to PEs and analyze their eigenvalue distributions. Fig.~\ref{fig:eigenvalues} plots the curves of accumulated energy/sum of top-$n$ eigenvalues versus the energy of all the eigenvalues. Supposing that we are using PCA to project the PEs of both matrices, to achieve the same explained variance ratio, our MJP needs more singular values (\emph{i.e.,} large dimensionality) than DeiT-S. This indicates that our positional embedding matrix is more informative. 
Similar observation is also revealed by Wang \emph{et al.}~\cite{wang2020position}.

\begin{figure*}[!t]
\centering
\includegraphics[width=.9\textwidth]{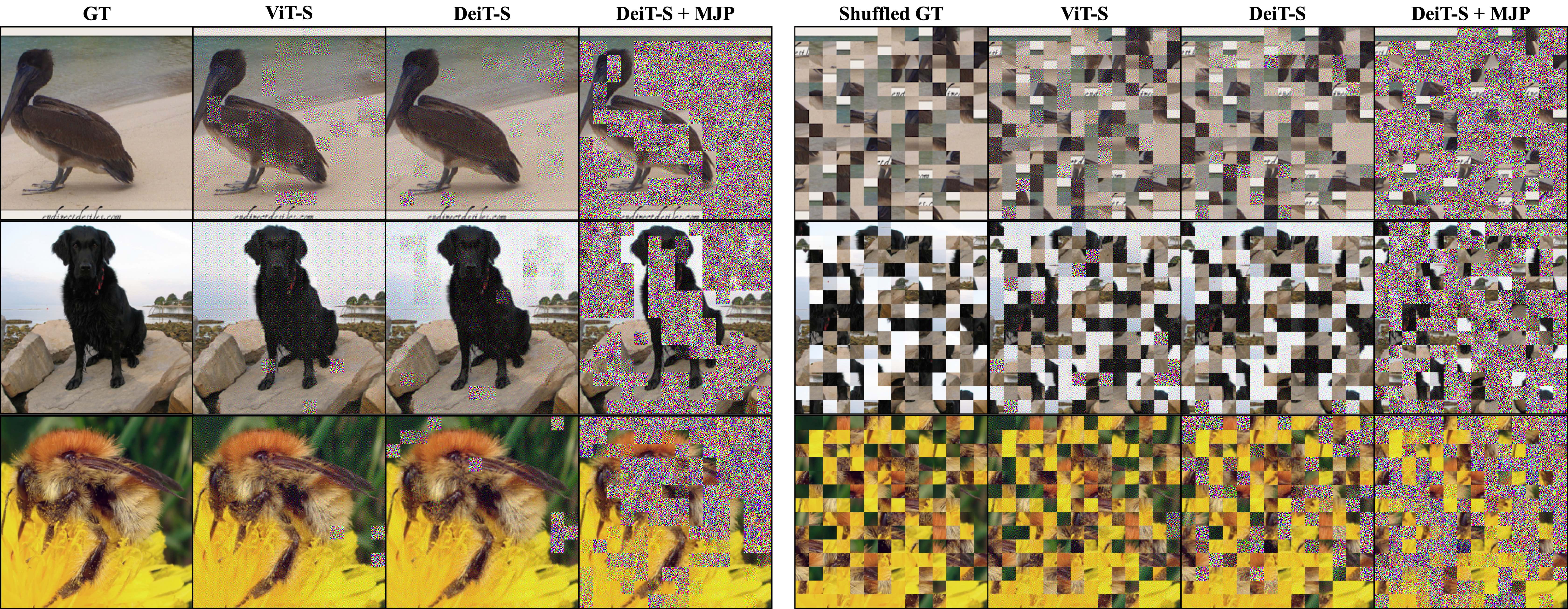}
\caption{Visual comparisons on image recovery with gradient updates~\cite{lu2022april}. Our proposed DeiT-S+MJP model significantly outperforms the original ViT-S~\cite{dosovitskiy2020image} and DeiT-S~\cite{touvron2021training} models.  
}
\label{fig:gradient-attack}
\vspace{-1em}
\end{figure*}

\subsection{Privacy Preservation}
The fundamental principle of the gradient attack methods in federated learning is that each sample activates only a portion of content-related neurons in the deep neural networks, leading to one specific backward gradients for one related samples (\emph{i.e.}, 1-to-1 mapping). Based on such an observation, we argue that feeding ViTs with input patches with permuted sequences may intuitively mislead the  attack. This is because now both the original and the transformed inputs may be matched to the same backward gradients (\emph{i.e.}, $n$-to-1 mapping).

To validate such an assumption, we utilize the public protocols\footnote{\url{https://github.com/JonasGeiping/breaching}} to recover image with gradient updates in the privacy attack. In this privacy attack, we apply the Analytic Attack proposed in APRIL~\cite{lu2022april}, which is designed for attacking the ViTs. We randomly sample 1K images from the validation set of ImageNet-1K (\emph{i.e.}, one image per category). To evaluate the anti-attack performance of a model, we introduce image similarity metrics to account for pixel-wise mismatch, including Mean Square Error (MSE), Peak Signal-to-Noise Ratio (PSNR), cosine similarity in the Fourier space (FFT$_\text{2D}$), and Learned Perceptual Image Patch Similarity (LPIPS)~\cite{zhang2018unreasonable}. Different from the evaluation in gradient attacks~\cite{lu2022april,hatamizadeh2022gradvit,yin2021see}, \emph{we suppose a model is with better capacity of privacy preservation when the recovered images from its gradient updates are less similar to the ground truth images}. 

\begin{table}[!t]
  \centering
  \caption{Comparisons on gradient leakage by analytic attack~\cite{lu2022april} with ImageNet-1K validation set, where we test (1) ViT-S, DeiT-S and our model in the setting (a); (2) ViT-S, DeiT-S and our model in the setting (b) (\emph{i.e.}, MJP with $\gamma=0.27$); (3) ablation on without (w/o) using $\mathbf{E}_{\text{unk}}$ in setting (a); and (4) Our model in setting (c).}
  \vspace{-1em}
  \resizebox{1.0\columnwidth}{!}{
  \setlength{\tabcolsep}{0.15em} 
  \begin{tabular}{XllcccccccX}
    \toprule
     & \textbf{Model} &  \textbf{Set.} & \textbf{Acc.} $\uparrow$  & \textbf{MSE} $\uparrow$ & \textbf{FFT$_\text{2D}$} $\uparrow$ & \textbf{PSNR}~$\downarrow$ & \textbf{SSIM} $\downarrow$ & \textbf{LPIPS} $\uparrow$ 
     \\ \midrule 
     \multirow{4}{*}{(1)} & ViT-S~\cite{dosovitskiy2020image} & \multirow{4}{*}{a} & 78.1 &  .0278 & .0039 & 19.27 & .5203 & .3623 \\
     & DeiT-S~\cite{touvron2021training} & & 79.8  & .0350 & .0057 & 18.94 & .5182 & .3767 \\
     & DeiT-S (w/o PEs) & & 77.5 & .0379 & .0082 & 20.22 & .5912 & .2692 \\
     & DeiT-S+MJP & & \textbf{80.5}  & \textbf{.1055}  & \textbf{.0166} & \textbf{11.52} & \textbf{.4053} & \textbf{.6545} 
     \\ \midrule
     \multirow{4}{*}{(2)} & ViT-S~\cite{dosovitskiy2020image} & \multirow{4}{*}{b} & 18.7 &   .0327 & .0016 & 18.44 & .6065 & .2836 \\
     & DeiT-S~\cite{touvron2021training} & & 36.0 & .0391 & .0024 & 17.60 & .5991 & .3355 \\ 
     & DeiT-S (w/o PEs) & & \textbf{77.5} & .0379 & .0025 & 20.25 & .6655 & .2370 \\
     & DeiT-S+MJP & & 62.9 &   \textbf{.1043} & \textbf{.0059} & \textbf{11.66} & \textbf{.4493} & \textbf{.6519} 
     \\ \midrule
     (3) & DeiT-S+MJP (w/o) & a  &  40.6 & .1043 & .0059 & 11.66 & .4493 & .6519 
     \\ \midrule
     (4) & DeiT-S+MJP & c & 62.9   &  .1706 & .0338 & 8.07 & .0875 & .8945 \\ \bottomrule
  \end{tabular}}
  \label{tab:grad-attack}
  \vspace{-1em}
\end{table}

Given an image $\mathbf{x}$ and its transformed (\ie, patch shuffled) version $\tilde{\mathbf{x}}$, a ViT model $\mathcal{M}$, and automatic evaluation metrics $\phi$, we conduct three different settings for fair comparisons: (a) $\phi(\nabla{\mathcal{M}(\mathbf{x})}, \mathbf{x})$, (b) $\phi(\nabla{\mathcal{M}(\tilde{\mathbf{x}})}, \tilde{\mathbf{x}})$, and (c) $\phi(\nabla{\mathcal{M}(\tilde{\mathbf{x}})}, \mathbf{x})$, where $\nabla$ refers to recovering input image through gradient attacks. Table~\ref{tab:grad-attack} shows the quantitative comparisons between our method and the original ViTs for batch gradient inversion on ImageNet-1K. APRIL~\cite{lu2022april} enables a viable, complete recovery of original images from the gradient updates of the original ViTs. However, it performs worse in recovering from ``DeiT-S+MJP'', leading to best performances on all evaluation metrics and outperform others by a large margin. 

More surprisingly, our proposed method makes APRIL yield unrecognizable images and fail in recovering the details in the original images (\emph{i.e.}, noisy patches in the outputs), as shown in Fig.~\ref{fig:gradient-attack}. The left four columns in Fig.~\ref{fig:gradient-attack} are tested on original images, where all PEs are standard and correspond to their patch embeddings. Meanwhile, the right four columns are tested with transformed ones, where the shuffled patches are with the shared unknown PEs. Both the visual and quantitative comparisons verify that our MJP alleviates the gradient leakage problem. We also notice that DeiT-S without using PEs is inclined to be at higher risk of privacy leakage (\emph{i.e.}, easier to be attacked by gradients). These promising results indicate that our MJP is a promising strategy to protect user privacy in federated learning. 

\begin{table}[t]
  \centering
  \caption{Explained variance versus PCA projected dimensionality. }
  \vspace{-2mm}
  \resizebox{1.0\columnwidth}{!}{
  \begin{tabular}{cccccc}
    \toprule
     \textbf{Projected Dimension} & 3 & 4 & 5 & 6 & 7 \\
    \midrule
    \textbf{Deit-S EV (\%)} &54.61 & 68.55 & 77.95 & 85.54& 90.74\\
    \textbf{DeiT-S+MJP EV (\%)} &46.74 &58.36 &69.10 &78.13 &84.55 \\
    \bottomrule
  \end{tabular}
  \label{tab:variance_dim}
  }
  \vspace{-2em}
\end{table}

\subsection{Discussion}
\noindent\textbf{PCA Projected Dimensionality.} Table~\ref{tab:variance_dim} presents the explained variance (EV) of our DeiT-S+MJP and Deit-S versus different projection dimension. A low dimensionality can explain a large amount of information, which proves that the embedding matrix is sparse in nature. Moreover, to achieve the same explained variance ratio, our DeiT-S+MJP needs a large dimensionality than Deit-S. This indicates that the positional embedding matrix of DeiT-S+MJP is less sparse but more informative.

\noindent\textbf{Accuracy \emph{Vs.} Shuffle Ratio.}
Intuitively, with the increase of the shuffle ratio from the proposed MJP method, the original intrinsic inductive bias will be undermined. However, from the experimental results, the performance of ViTs is actually boosted via the proposed MJP. To figure out the reason behind such counter-intuitive phenomenon, we visualize the last self-attention of our proposed method in Fig.~\ref{fig:attention} of our Sipp. Mat. It shows that the attention heads of our method present more diverse and content-aware attentions than original DeiT-S. We think that it might be attributed to the proposed DAL, which strengthens the spatial information of the unmasked PEs in a more efficient manner. To this end, such learned content-aware attention becomes more meaningful and results in the better accuracy.

%% file: latex/sections/6conclusion.tex
\section{Conclusion}
\label{sec:conclusion}
In this paper, we first visually demonstrate that PEs can explicitly learn the 2D spatial relationship from the input patch sequences. By feeding ViTs with transformed input, we identify the issue that PEs may weaken the position-insensitive property. Based on this observation, we propose an easy-to-reproduce yet effective Masked Jigsaw Puzzle (MJP) position embedding method to alleviate the conflict in PEs (preserving the consistency \emph{versus} maintaining the accuracy). Experimental results show the proposed MJP can bring the the position-insensitive property back to ViTs without degenerating the accuracy on the large-scale dataset (\emph{i.e.}, ImageNet-1K). In a certain sense, the proposed MJP is also a data augmentation technique, which boosts the robustness to the common corruptions (\emph{e.g.}, ImageNet-C) and adversarial examples (\emph{e.g.}, ImageNet-A/O). Surprisingly, MJP can improve the privacy preservation capacity of ViTs under typical gradient attacks by a large margin, which may pilot a new direction for privacy preservation.

%% file: latex/sections/7appendix.tex
\appendix

\section{Training Settings}
In the experiments on ImageNet-1K, we employ an AdamW~\cite{kingma2015adam} optimizer for 300 epochs using a cosine decay learning rate scheduler and 20 epochs of linear warm-up. A batch size of 1024, an initial learning rate of 0.001, and a weight decay of 0.05 are used. We include most of the augmentation and regularization strategies (e.g., repeated augmentation~\cite{hoffer2020augment}, CutMix~\cite{yun2019cutmix}, and Mixup~\cite{zhang2018mixup}) of~\cite{touvron2021training} in training, as shown in Table~\ref{tab:comp_hyperparameters}.

\begin{table}[!ht]
\centering
\caption{
 Ingredients and hyper-parameters for our method.
\label{tab:comp_hyperparameters}}
\begin{tabular}{lccc}
\toprule
Epochs   &  300     \\
\midrule
Batch size & 1024\\
Optimizer &  AdamW\\
     learning rate       &  $0.0005\times \frac{\textrm{batchsize}}{512} $  \\
     Learning rate decay &  cosine  \\
     Weight decay        &  0.05    \\
     Warmup epochs  &  20       \\
\midrule
     Label smoothing $\epsilon$ & 0.1     \\
     Dropout      &  $\times$     \\
     Repeated Aug &  $\surd$ \\
     Gradient Clip. & $\surd$ \\
\midrule
     Rand Augment  &  9/0.5 \\
     Mixup prob.  &  0.8     \\
     Cutmix prob.   &  1.0    \\
     Erasing prob.    & 0.25    \\
 \bottomrule
\end{tabular}
\end{table}

\section{Training Efficiency} 
\begin{figure}[t]
     \centering
     \begin{subfigure}[b]{0.3\textwidth}
         \centering
         \includegraphics[width=\textwidth]{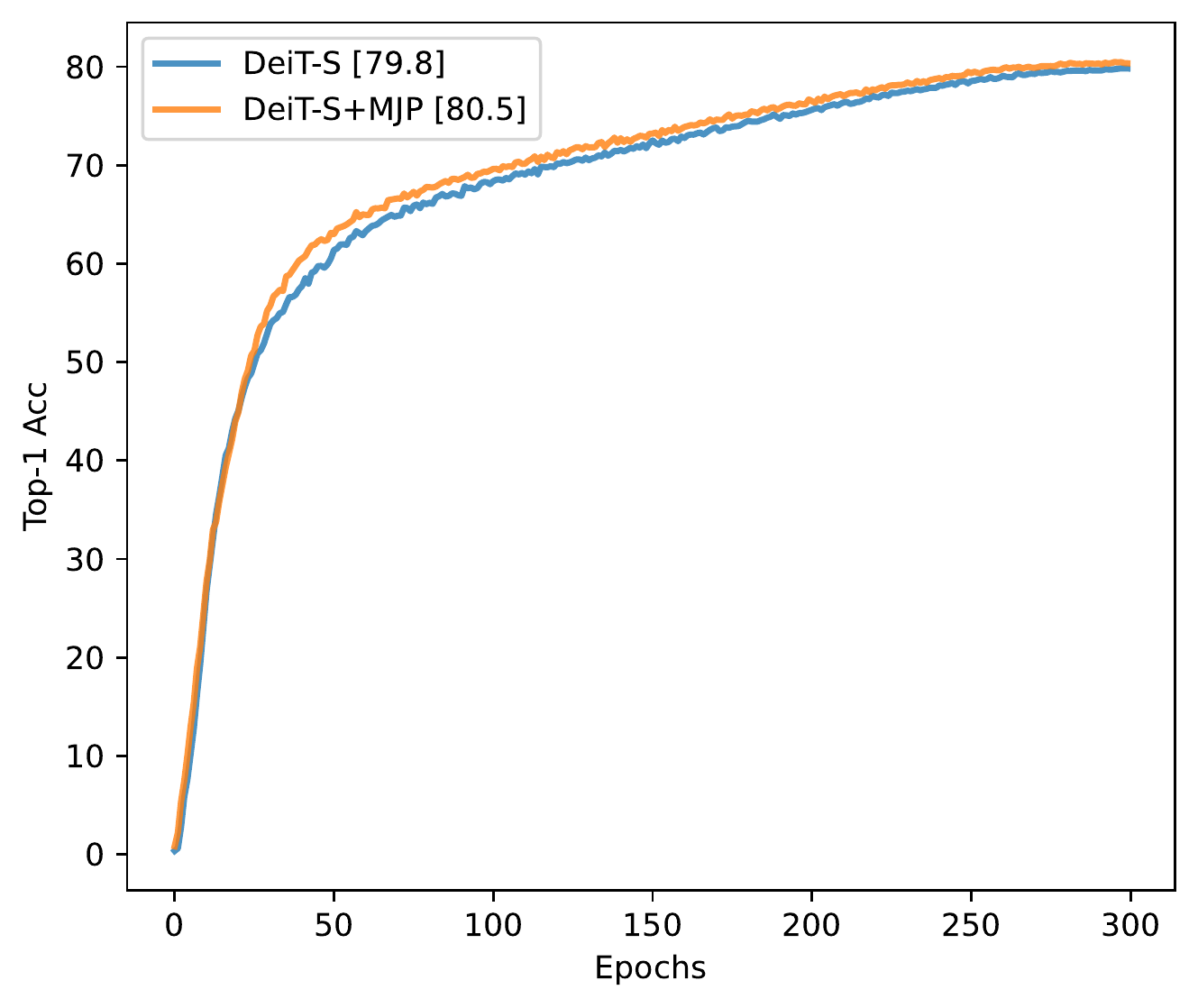}
         \caption{}
     \end{subfigure}
     \hfill
     \begin{subfigure}[b]{0.3\textwidth}
         \centering
         \includegraphics[width=\textwidth]{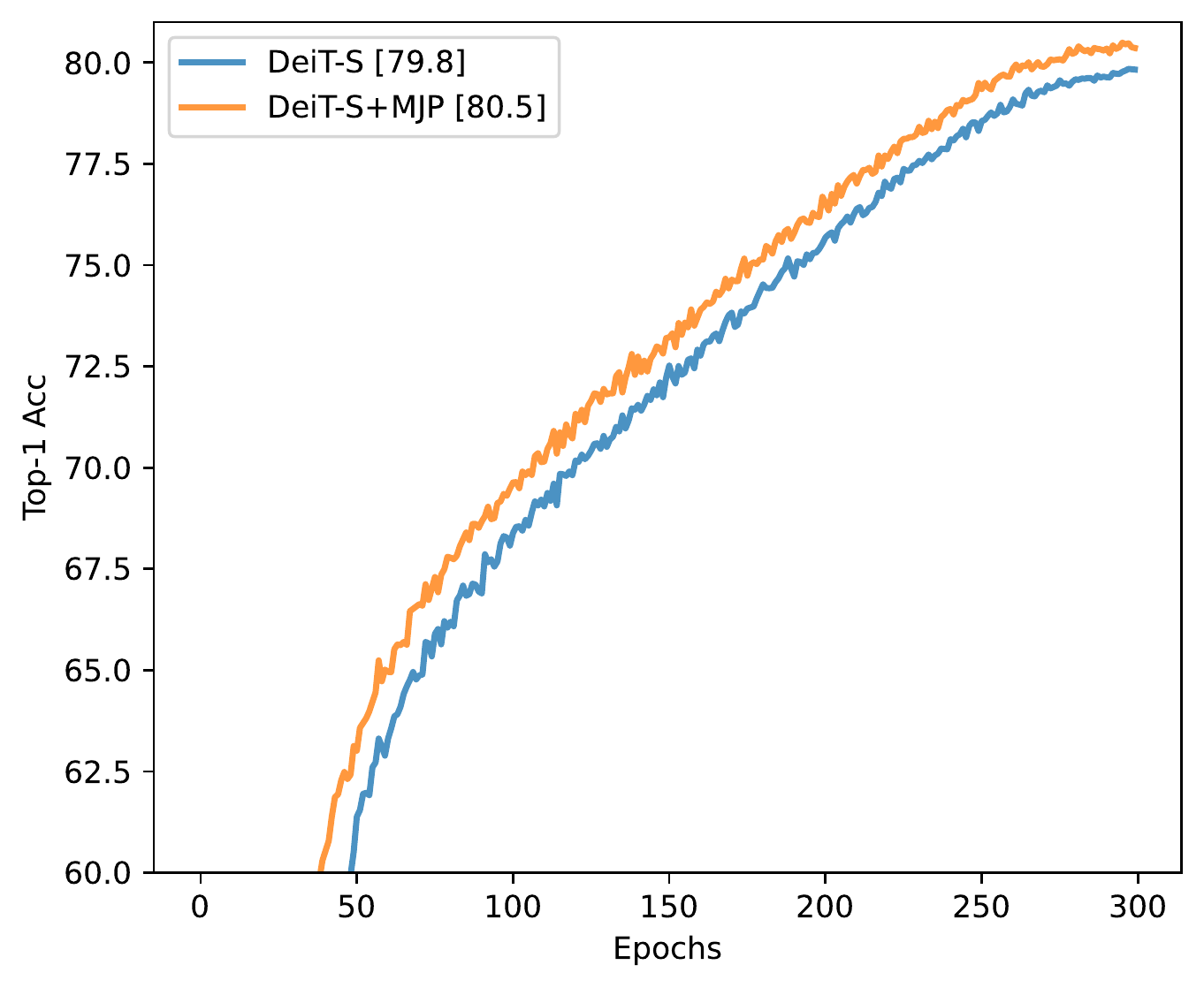}
         \caption{}
     \end{subfigure}
        \caption{Comparisons the top-1 accuracy of DeiT-S and DeiT-S+MJP during the training: (a) the whole training and (b) a zoom-in screenshot for the accuracy larger than 60\%.}
        \label{fig:training-efficiency}
\end{figure}

We train our proposed method on ImageNet-1K with 8 V100 NVIDIA GPUs. We note that the computational consumption of the MJP procedure is negligible (\emph{i.e.}, $+2\%$ time consumption per epoch). Meanwhile, MJP  accelerates the convergence during the training as show in Figure~\ref{fig:training-efficiency}. 

\section{Position Embeddings}

\begin{figure*}[t]
     \centering
     \begin{subfigure}[b]{0.32\textwidth}
         \centering
         \includegraphics[width=1\linewidth]{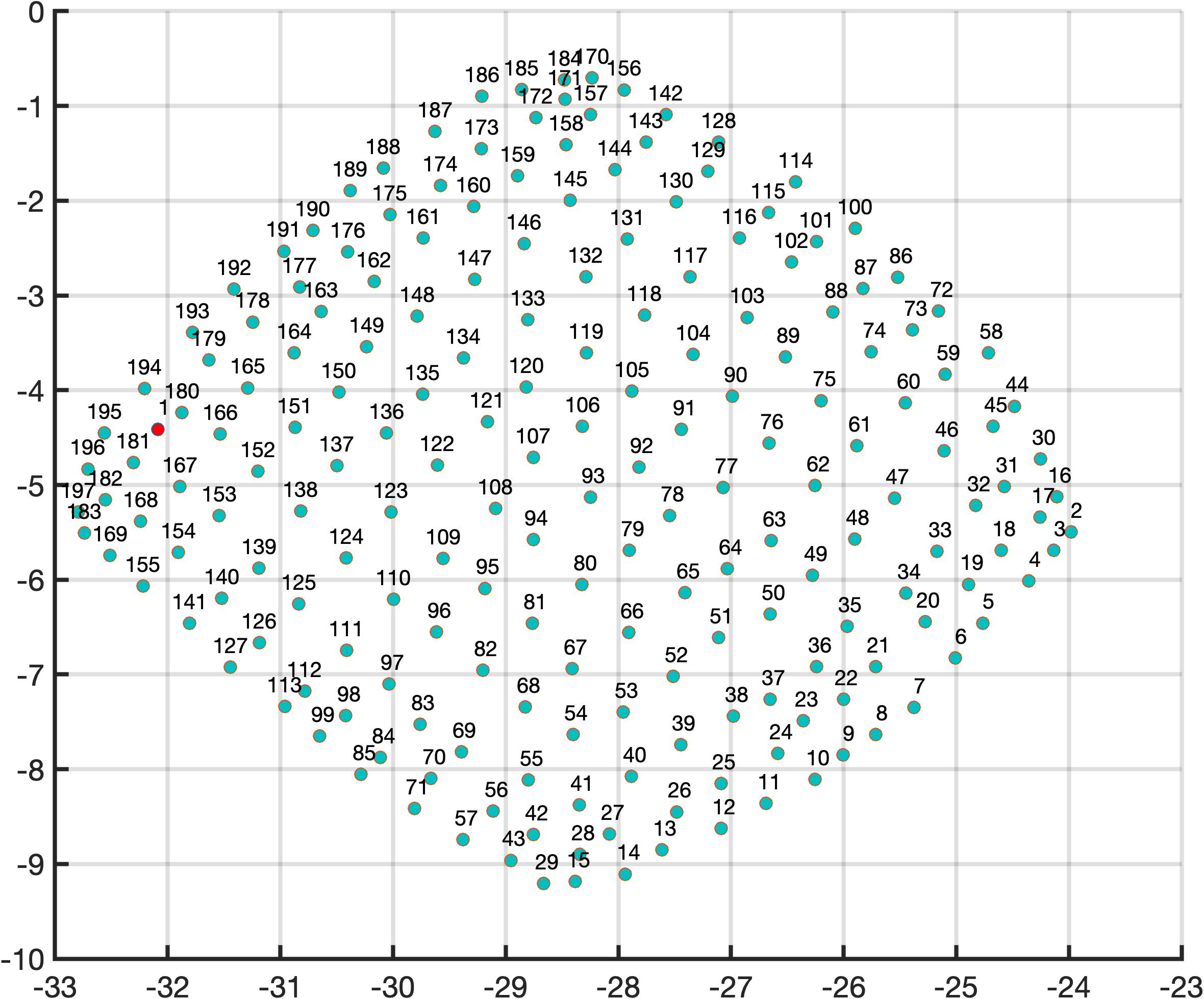}
         \caption{}
     \end{subfigure}
     \hfill
     \begin{subfigure}[b]{0.32\textwidth}
         \centering
         \includegraphics[width=1\linewidth]{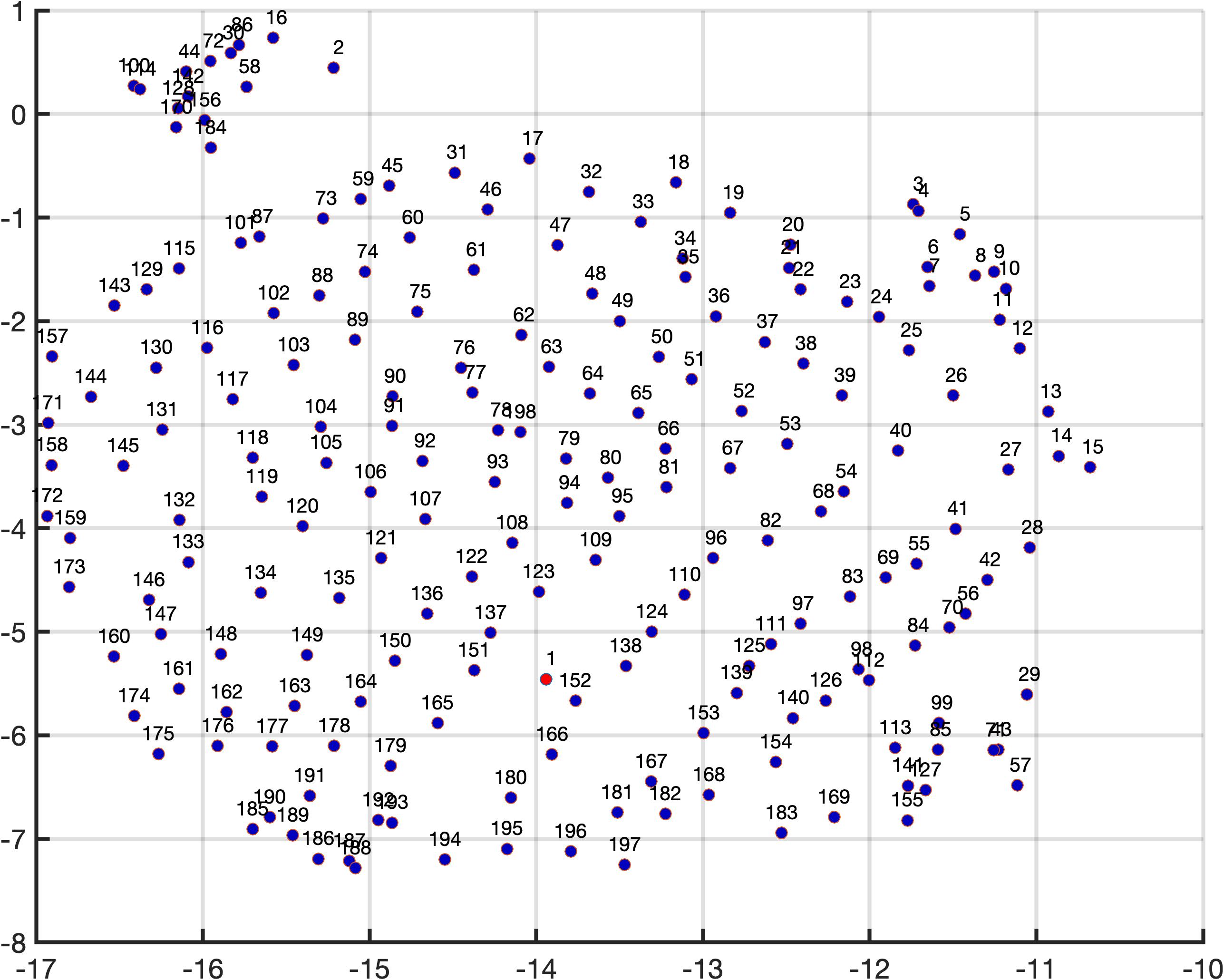}
         \caption{}
     \end{subfigure}
     \hfill
     \begin{subfigure}[b]{0.32\textwidth}
         \centering
         \includegraphics[width=1\linewidth]{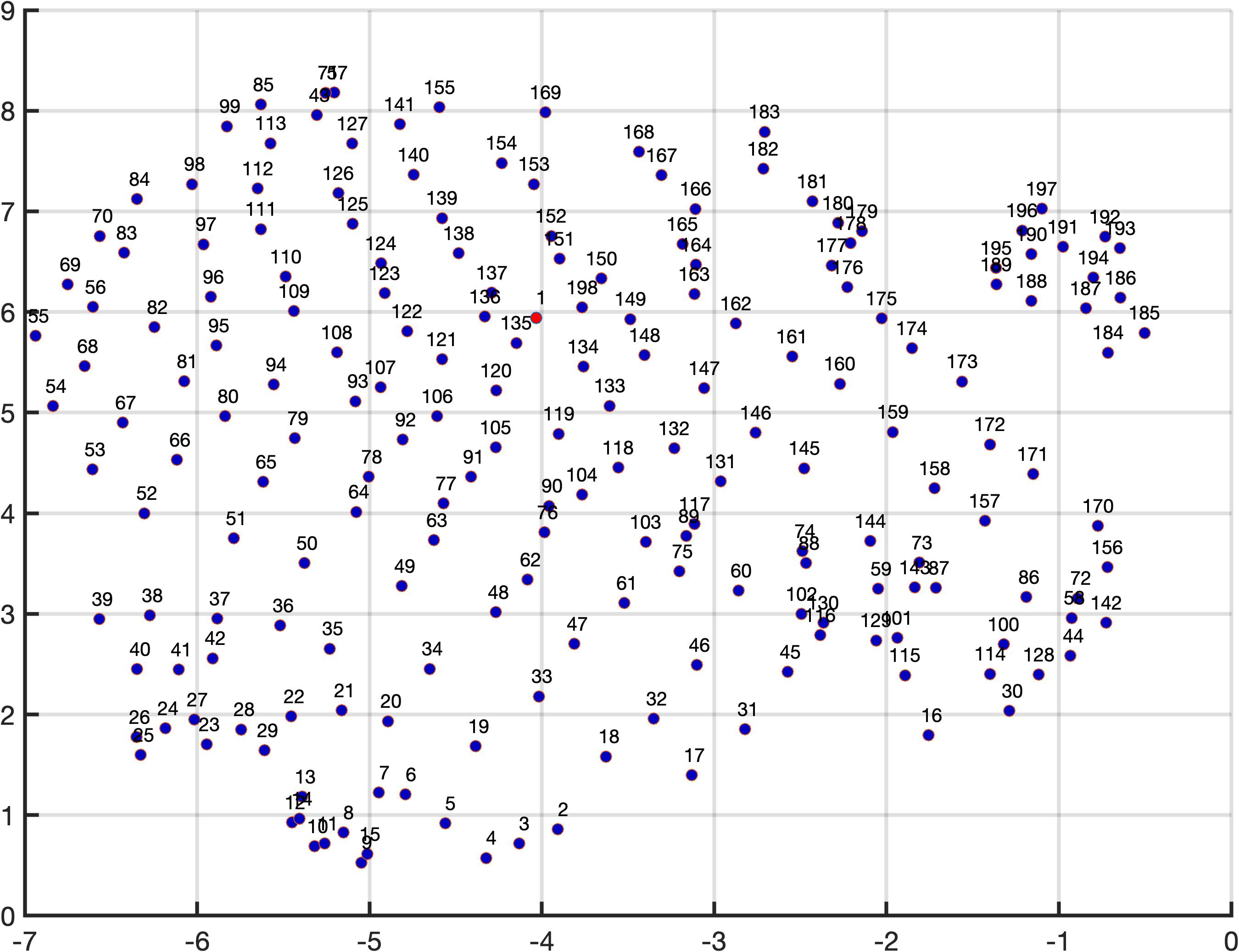}
         \caption{}
     \end{subfigure}
     \hfill
     \begin{subfigure}[b]{0.32\textwidth}
         \centering
         \includegraphics[width=1\linewidth]{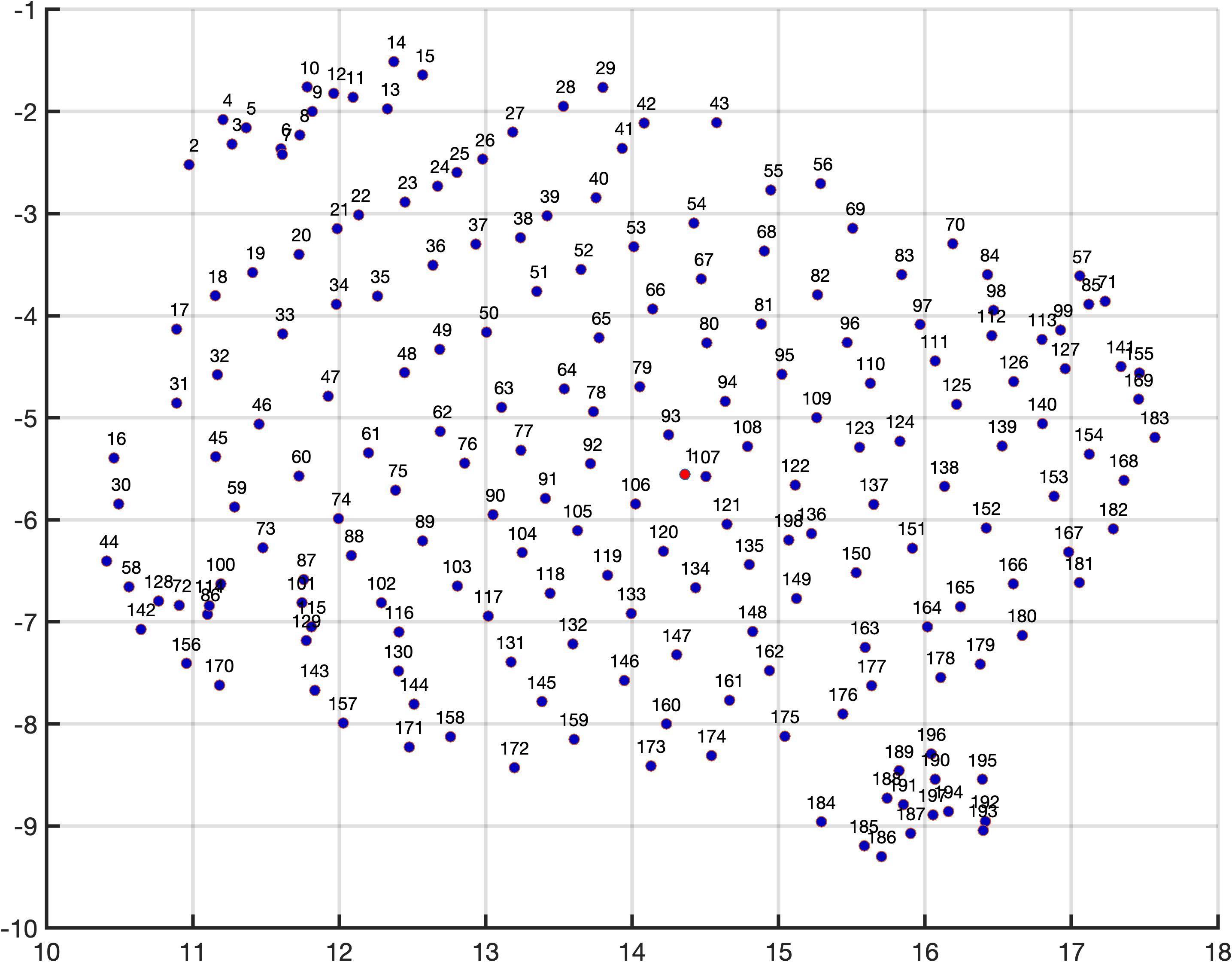}
         \caption{}
     \end{subfigure}
     \hfill
     \begin{subfigure}[b]{0.32\textwidth}
         \centering
         \includegraphics[width=1\textwidth]{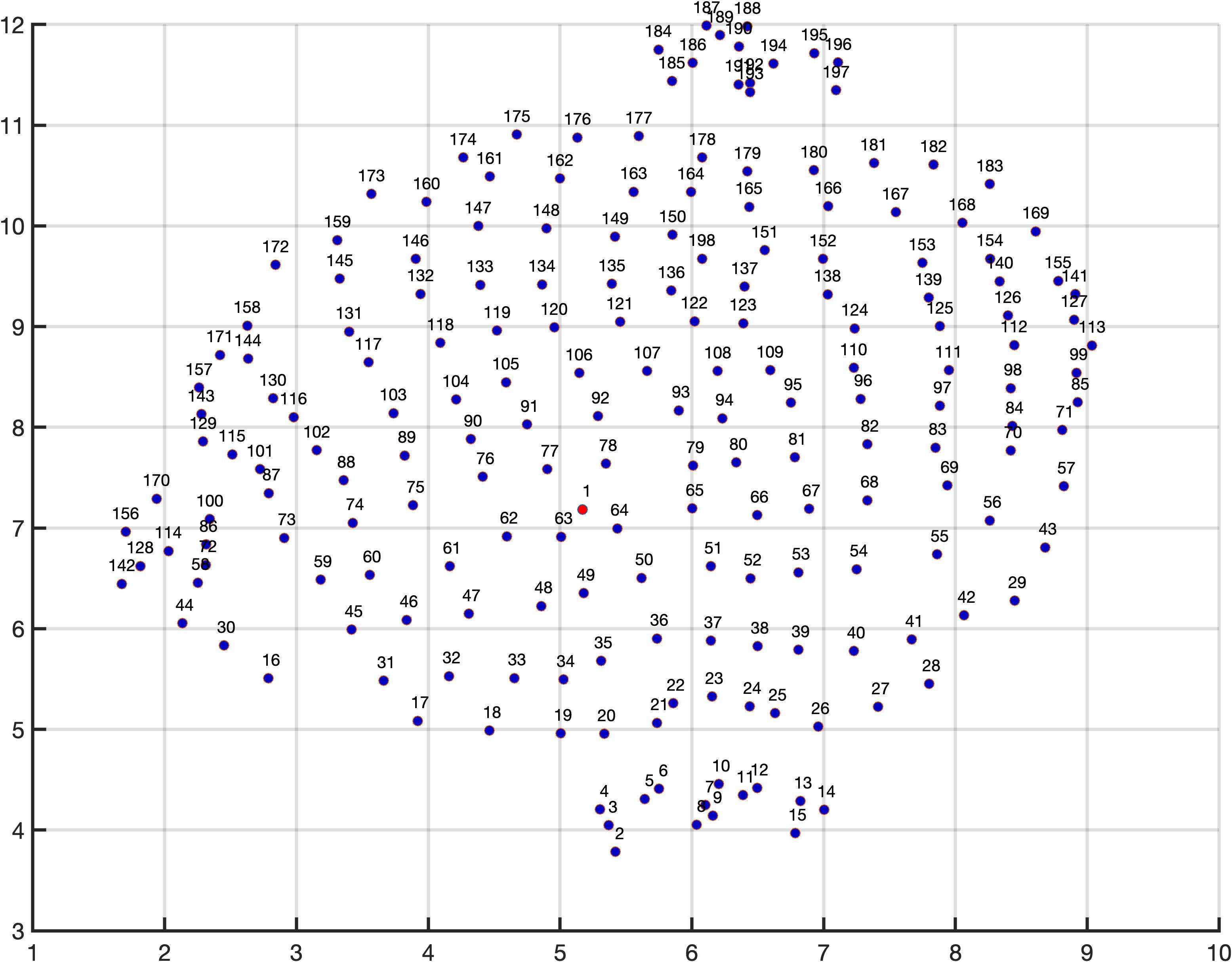}
         \caption{}
     \end{subfigure}
     \hfill
     \begin{subfigure}[b]{0.32\textwidth}
         \centering
         \includegraphics[width=1\linewidth]{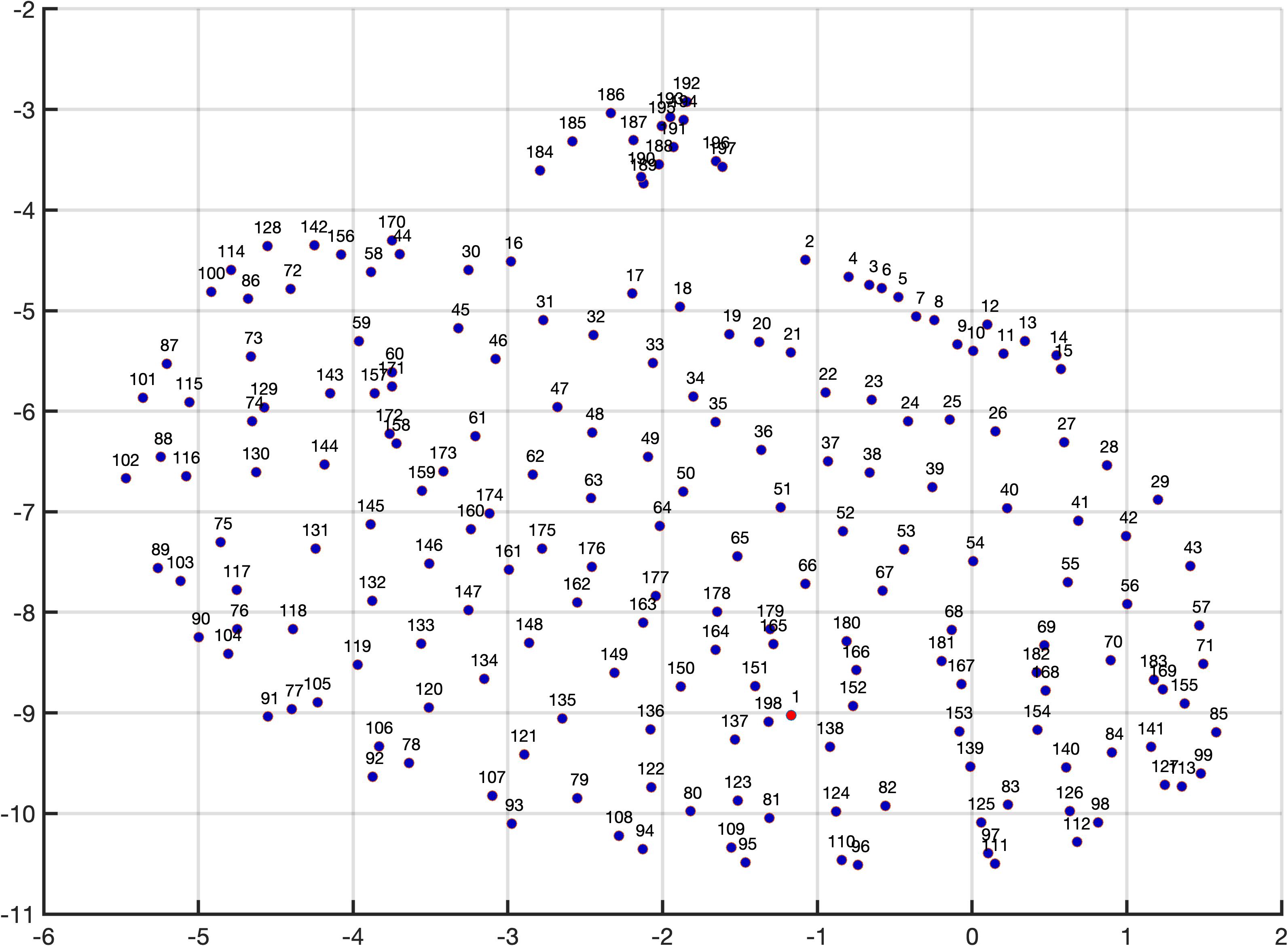}
         \caption{}
     \end{subfigure}
    \caption{UMAP projections of the position embeddings collected from: (a) the original DeiT-S~\cite{dosovitskiy2020image}, (b) - (f) are DeiT-S+MJP trained with masking ratio $\gamma = \{0.03, 0.09, 0.15, 0.21, 0.27\}$ respectively.}
    \label{fig:umap-projection}
\end{figure*}

\subsection{Position Regression}
Inspired by Wang \emph{et al.}~\cite{wang2020position}, we also check whether a position embedding can actually capture its absolute position. To some extent, such position information could be reconstructed by a reversed mapping function $g: \mathcal{X} \to \mathcal{P}$, where $\mathcal{X}$ and $\mathcal{P}$ are embedding space and position space, respectively. Thus, we use linear regression to learn such a function $g$ that transfers the embeddings to the original positions. For the position embeddings in ViTs, we can map them into either 1-D sequence space or 2-D patch grid space. Given we only have 196 data points (\emph{i.e.}, 224$\times$224 image resolution with 16$\times$16 patch size) for each learned embedding, a 5-fold cross-validation is applied to avoid overfitting. The reversed mapping functions are evaluated by Mean Absolute Error (MAE), and the result is shown in Table~\ref{tab:abs-pos-reg}. 

\begin{table}[t]
  \centering
  \caption{Mean absolute error of the reversed mapping function learned by linear regression.}
  \resizebox{1.0\columnwidth}{!}{
  \begin{tabular}{lrr}
    \toprule
     \textbf{Method} & \textbf{1D MAE} & \textbf{2D MAE} \\ \midrule
     DeiT-S~\cite{touvron2021training} & .945$\pm_{.031}$ & .076$\pm_{.004}$ \\
     DeiT-S + MJP (DAL - LN) & .566$\pm_{.013}$ & .042$\pm_{.002}$ \\
     DeiT-S + MJP (DAL - NLN) & 1.301$\pm_{.035}$ & .134$\pm_{.003}$ \\
    \bottomrule
  \end{tabular}
  \label{tab:abs-pos-reg}
  }
\end{table}

\begin{table}[t]
  \centering
  \caption{Segmentaion results on ADE20K dataset (Pre-trained on ImageNet-1K).}
  \begin{center}
  \end{center}
  \vspace{-3em}
  \resizebox{1.0\columnwidth}{!}{
  \begin{tabular}{lrrr}
    \toprule
    \centering
     \textbf{Method} & \textbf{Top-1 Acc} & \textbf{mIoU} & \textbf{mAcc}\\ \midrule
     Swin-Tiny~\cite{liu2021swin_rebuttal} & 81.3 & 43.87 & 55.22 \\
     Swin-Tiny + MJP & 81.3 & 44.03 (+0.16) & 55.50 (+0.28) \\
    \bottomrule
  \end{tabular}
  \label{tab:segmentation}
  }
  \vspace{-0.8em}
\end{table}

From the results, the reversed mapping function of learnt position embeddings by ``DeiT-S + MJP (DAL-LN)'' can better represent the absolute positions. Meanwhile, the embeddings learned by the original DeiT-S and ``DeiT-S + MJP (DAL-NLN)'' also well learn the information about the absolute positions. Similar to~\cite{wang2020position}, we have also tried some more complicated non-linear models such as SVM or MLP to map the embeddings back, which suffer from overfitting issue and perform worse. This implies that the position information in ViTs can actually be modeled by a linear model, which is consistent with Transformer encoders used in NLP field. Besides, the MAE of ``DeiT-S+MJP (DAL - NLN)'' is larger than the MAE of ``DeiT-S+MJP (DAL - LN)'' in Table ~\ref{tab:abs-pos-reg}. It indicates the nonlinear regression during the training aggregate more information beyond the position information (\emph{i.e.}, more informative). The results are consistent with the Fig.4 in our main paper.

\subsection{Projections of Position Embeddings}
As shown in Figure~\ref{fig:umap-projection}, the position embeddings learned by the original DeiT-S is in a form of structured grids. Once we introduce MJP strategy to the training, it makes the projection of these position embeddings less structured. Meanwhile, the spatial-wise relative position information is preserved. We assume that the additional information (\emph{i.e.}, more informative) in the position embeddings leads to such a difference.

\section{Fine-tuning for Semantic Segmentation}
A fine-tuning experiment on ADE20K dataset for semantic segmentation is presented. We use the popular UperNet architecture with a Swin-Tiny backbone pre-trained on ImageNet-1K. The re-
sults shown in Tab.~\ref{tab:segmentation} indicate that our MJP doesn’t have
negative effects on other regular position-sensitive tasks.

\begin{table*}[!t]
  \centering
  \caption{Comparisons on robustness to common corruptions with ImageNet-C.}
  \begin{tabular}{lcccccccccccc}
    \toprule
     \textbf{Method} & \rotatebox{90}{Gaussian noise} & \rotatebox{90}{Shot noise} & \rotatebox{90}{Impulse noise} & \rotatebox{90}{Defocus blur}  & \rotatebox{90}{Glass blur} & \rotatebox{90}{Motion blur} & \rotatebox{90}{Contrast} & \rotatebox{90}{Elastic} & \rotatebox{90}{Pixelate} & \rotatebox{90}{JPEG} & \textbf{mCE} $\downarrow$ \\ \midrule
     DeiT-S & 41.0 & 42.7 & 42.8 & 50.5 & 59.4 & \textbf{45.5} & 36.1 & \textbf{43.0} & 42.4 & 36.6 & 44.0 \\
     \rowcolor{Light}DeiT-S + MJP & \textbf{39.9} & \textbf{41.8} & \textbf{41.2} & \textbf{49.9} & \textbf{58.6} & 47.4 & \textbf{34.5} & 43.3 & \textbf{40.0} & \textbf{35.8} & \textbf{41.5} \\
    \bottomrule
  \end{tabular}
  \label{tab:supp-ablation-corruptions}
\end{table*}

\section{Robustness to Corruptions}
We show the details on the evaluation with ImageNet-C~\cite{hendrycks2019benchmarking}, as shown in Table~\ref{tab:supp-ablation-corruptions}. Compared to the original DeiT-S, our method achieves better performance on most tested corruptions\footnote{In the camera-ready version, we reported the unnormalized results. However, in accordance with the conventions of the robustness literature, we have updated this version to include the normalized values. Nonetheless, it is important to note that our MJP consistently outperforms the DeiT-S model for both evaluation methods.}.

\section{Visualization of the last self-attention}
\begin{figure}[t]
    \centering
    \includegraphics[width=.93\linewidth]{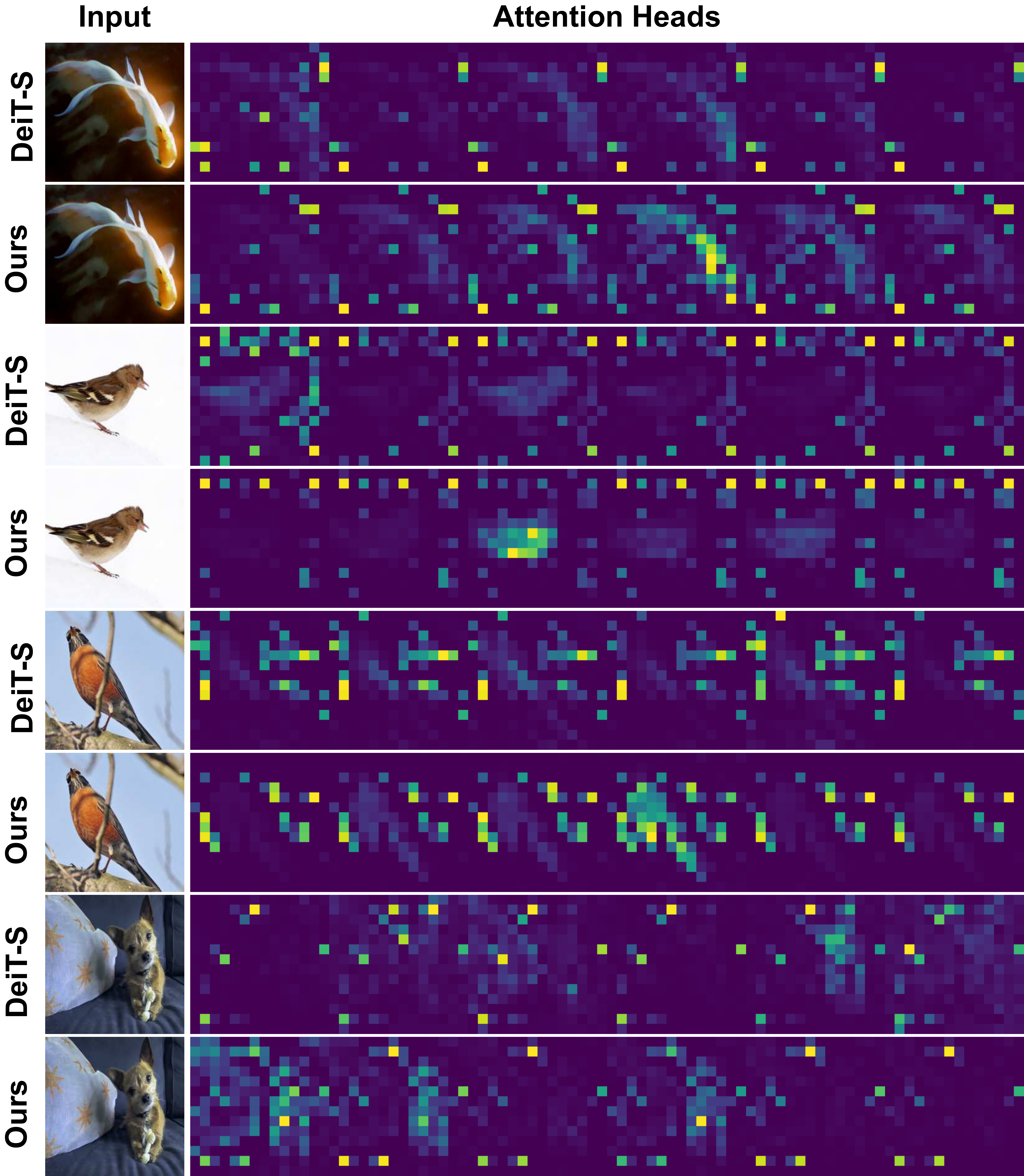}
    \caption{Visual comparisons between visualization maps of the last self-attention in DeiT-S~\cite{touvron2021training} and our proposed DeiT-S+MJP.}
    \label{fig:attention}
    \vspace{-1em}
\end{figure}

The visualization the last self-attention of our method in Fig.~\ref{fig:attention}. It shows that the attention heads of our method present more diverse and content-aware attentions than original DeiT-S.

\section{More details on Privacy Preserving}
\subsection{Privacy Protection by Random Patch Permutation}
Existing analytic gradient attack algorithms mainly model the problem as a linear system with closed-form solutions~\cite{zhu2020r,lu2022april}. For ViTs, the linear system is defined as:
\begin{equation}
    \frac{\partial l }{\partial \mathbf{z}_{0}}\mathbf{z}_{0}^{T}  = \pmb{Q}^{T}\frac{\partial l}{\partial\pmb{Q}} + \pmb{K}^{T}\frac{\partial l}{\partial\pmb{K}} + \pmb{V}^{T}\frac{\partial l}{\partial\pmb{V}}
\end{equation}
where $\mathbf{z}_{0}$ denotes the image embedding that consists of patch embedding and positional embedding (i.e., $\mathbf{z}_{0}{=}\pmb{x}_p\mathbf{E}{+}\mathbf{E}_{\text{pos}}$, where $\pmb{x}_p$ denotes the sequence of flattened 2D patches and $\mathbf{E}$ represents the trainable linear projection). Since we have $\nicefrac{\partial l}{\partial \mathbf{z}}{=}\nicefrac{\partial l}{\partial \mathbf{E}_{\text{pos}}}$, the positional embedding layer is thus vulnerable to the gradient leakage attack. When the gradient $\nicefrac{\partial l}{\partial \mathbf{E}_{\text{pos}}}$ is accessible, the image can be reconstructed as:
\begin{equation}
\label{eq:attack}
    \pmb{x}_p = \Big((\frac{\partial l }{\partial \mathbf{E}_{\text{pos}}})^{-1} (\pmb{Q}^{T}\frac{\partial l}{\partial\pmb{Q}} + \pmb{K}^{T}\frac{\partial l}{\partial\pmb{K}} + \pmb{V}^{T}\frac{\partial l}{\partial\pmb{V}}) - \mathbf{E}_{\text{pos}}\Big) \mathbf{E}^{-1}
\end{equation}
As indicated above, the gradient leakage of the PEs make the image easily reconstructed with closed-form solutions. To resolve this issue, we propose to randomly permute a portion of the image patches via our proposed block-wise random jigsaw shuffle algorithm \textbf{A.1}. The random permutation will drastically change both $\mathbf{E}_{\text{pos}}$ and $\nicefrac{\partial l }{\partial \mathbf{E}_{\text{pos}}}$ in the above equation. 
This could significantly increase the difficulties to solve the linear system and reconstruct the image.

\subsection{More visual results}
\begin{figure*}[!ht]
    \centering
    \includegraphics[width=0.97\linewidth]{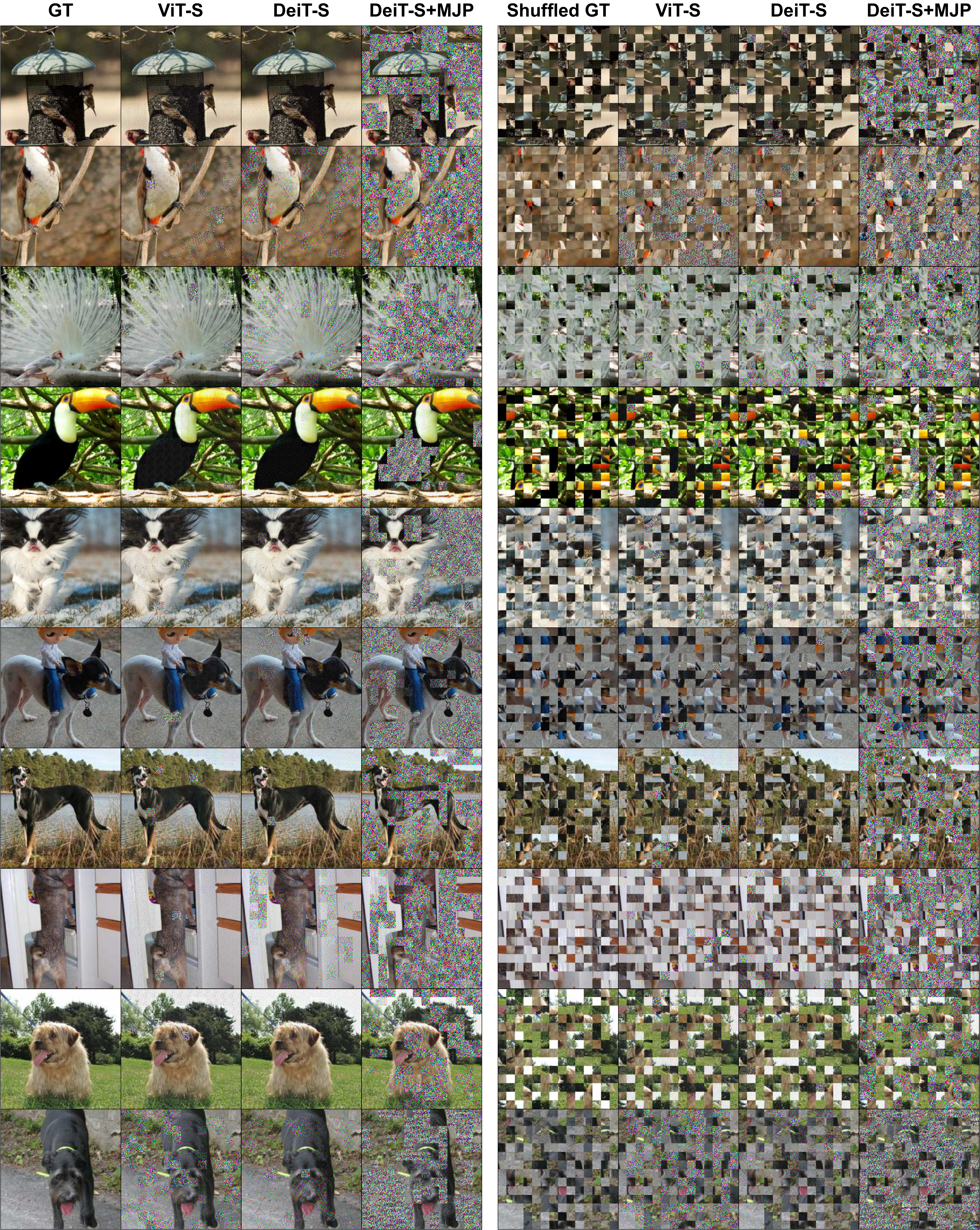}
    \vspace{-1em}
    \caption{Visual comparisons on image recovery with gradient updates~\cite{lu2022april}. We test both the original images without shuffling the patches and images shuffled with a masked ratio $\gamma=0.9$.}
    \label{fig:appendix-1}
\end{figure*}

\begin{figure*}[!ht]
    \centering
    \includegraphics[width=0.97\linewidth]{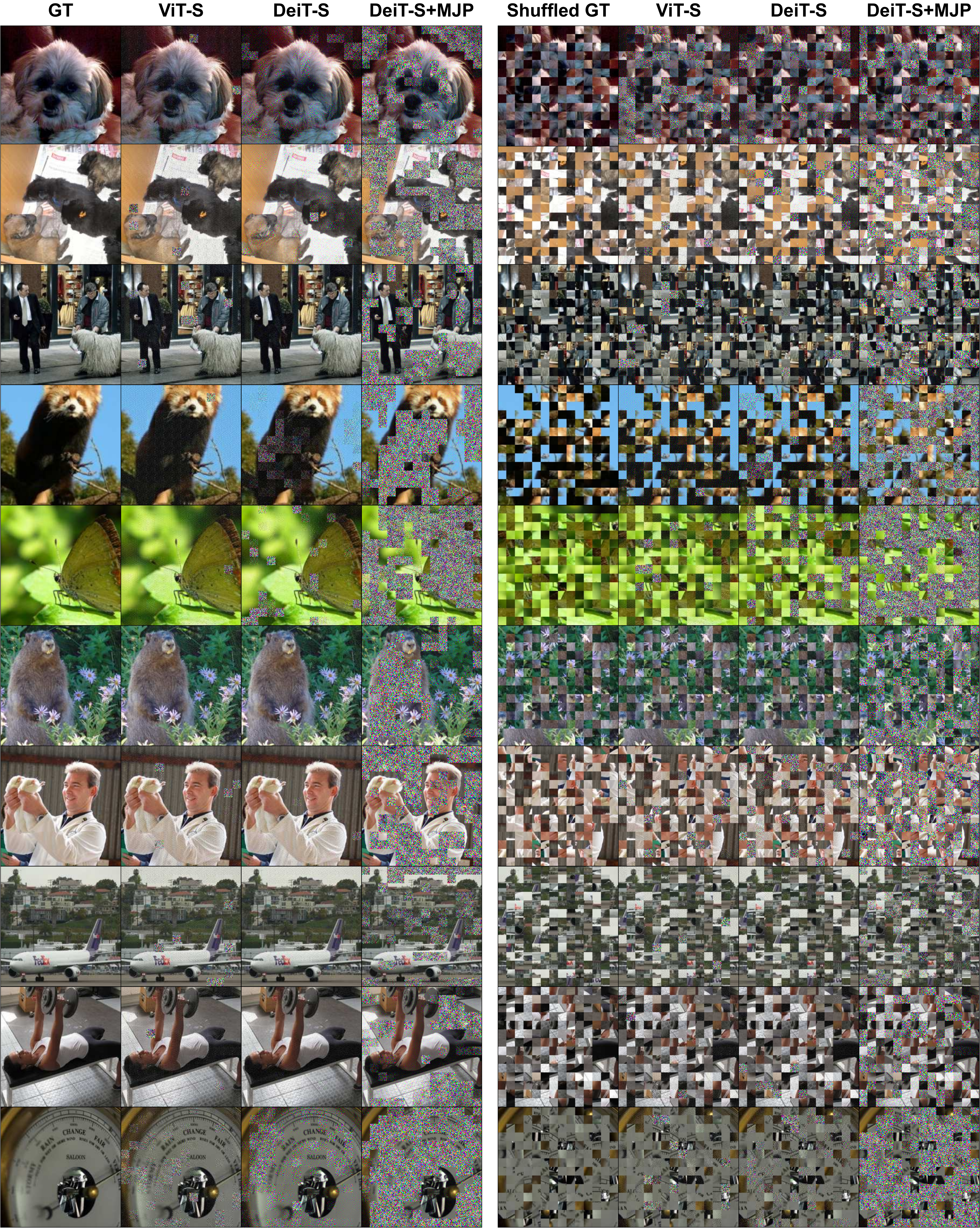}
    \vspace{-1em}
    \caption{Visual comparisons on image recovery with gradient updates~\cite{lu2022april}. We test both the original images without shuffling the patches and images shuffled with a masked ratio $\gamma=0.9$.}
    \label{fig:appendix-2}
\end{figure*}

Figure~\ref{fig:appendix-1} and Figure~\ref{fig:appendix-2} shows more visual results on image recovery with the gradient updates. Both these to figures clearly show that our method alleviates the privacy leakage issue a lot compared to the baselines, where most of the details are lost.

%% file: arXiv.bbl
\begin{thebibliography}{10}\itemsep=-1pt

\bibitem{liu2021swin_rebuttal}
Swin transformer: Hierarchical vision transformer using shifted windows.
\newblock In {\em ICCV}, 2021.

\bibitem{bao2021beit}
Hangbo Bao, Li Dong, and Furu Wei.
\newblock Beit: Bert pre-training of image transformers.
\newblock In {\em International Conference on Learning Representations (ICLR)},
  2022.

\bibitem{carion2020end}
Nicolas Carion, Francisco Massa, Gabriel Synnaeve, Nicolas Usunier, Alexander
  Kirillov, and Sergey Zagoruyko.
\newblock End-to-end object detection with transformers.
\newblock In {\em European Conference on Computer Vision (ECCV)}, 2020.

\bibitem{chen2021pre}
Hanting Chen, Yunhe Wang, Tianyu Guo, Chang Xu, Yiping Deng, Zhenhua Liu, Siwei
  Ma, Chunjing Xu, Chao Xu, and Wen Gao.
\newblock Pre-trained image processing transformer.
\newblock In {\em Proceedings of the IEEE/CVF Conference on Computer Vision and
  Pattern Recognition (CVPR)}, 2021.

\bibitem{chen2020generative}
Mark Chen, Alec Radford, Rewon Child, Jeffrey Wu, Heewoo Jun, David Luan, and
  Ilya Sutskever.
\newblock Generative pretraining from pixels.
\newblock In {\em International Conference on Machine Learning (ICML)}, 2020.

\bibitem{chu2021conditional}
Xiangxiang Chu, Zhi Tian, Bo Zhang, Xinlong Wang, Xiaolin Wei, Huaxia Xia, and
  Chunhua Shen.
\newblock Conditional positional encodings for vision transformers.
\newblock {\em arXiv preprint arXiv:2102.10882}, 2021.

\bibitem{devlin2018bert}
Jacob Devlin, Ming-Wei Chang, Kenton Lee, and Kristina Toutanova.
\newblock Bert: Pre-training of deep bidirectional transformers for language
  understanding.
\newblock In {\em Annual Conference of the North American Chapter of the
  Association for Computational Linguistics (NAACL)}, 2018.

\bibitem{dosovitskiy2020image}
Alexey Dosovitskiy, Lucas Beyer, Alexander Kolesnikov, Dirk Weissenborn,
  Xiaohua Zhai, Thomas Unterthiner, Mostafa Dehghani, Matthias Minderer, Georg
  Heigold, Sylvain Gelly, et~al.
\newblock An image is worth 16x16 words: Transformers for image recognition at
  scale.
\newblock In {\em International Conference on Learning Representations (ICLR)},
  2020.

\bibitem{dufter2022position}
Philipp Dufter, Martin Schmitt, and Hinrich Sch{\"u}tze.
\newblock Position information in transformers: An overview.
\newblock {\em Computational Linguistics}, 48(3):733--763, 2022.

\bibitem{fang2021you}
Yuxin Fang, Bencheng Liao, Xinggang Wang, Jiemin Fang, Jiyang Qi, Rui Wu,
  Jianwei Niu, and Wenyu Liu.
\newblock You only look at one sequence: Rethinking transformer in vision
  through object detection.
\newblock {\em Advances in Neural Information Processing Systems},
  34:26183--26197, 2021.

\bibitem{gehring2017convolutional}
Jonas Gehring, Michael Auli, David Grangier, Denis Yarats, and Yann~N Dauphin.
\newblock Convolutional sequence to sequence learning.
\newblock In {\em International Conference on Machine Learning (ICML)}, 2017.

\bibitem{gidaris2018unsupervised}
Spyros Gidaris, Praveer Singh, and Nikos Komodakis.
\newblock Unsupervised representation learning by predicting image rotations.
\newblock In {\em International Conference on Learning Representations}, 2018.

\bibitem{girdhar2019video}
Rohit Girdhar, Joao Carreira, Carl Doersch, and Andrew Zisserman.
\newblock Video action transformer network.
\newblock In {\em Proceedings of the IEEE/CVF conference on computer vision and
  pattern recognition}, pages 244--253, 2019.

\bibitem{han2020survey}
Kai Han, Yunhe Wang, Hanting Chen, Xinghao Chen, Jianyuan Guo, Zhenhua Liu,
  Yehui Tang, An Xiao, Chunjing Xu, Yixing Xu, et~al.
\newblock A survey on visual transformer.
\newblock {\em arXiv}, 2020.

\bibitem{hatamizadeh2022gradvit}
Ali Hatamizadeh, Hongxu Yin, Holger Roth, Wenqi Li, Jan Kautz, Daguang Xu, and
  Pavlo Molchanov.
\newblock Gradvit: Gradient inversion of vision transformers.
\newblock In {\em Proceedings of the IEEE/CVF Conference on Computer Vision and
  Pattern Recognition (CVPR)}, 2022.

\bibitem{he2022masked}
Kaiming He, Xinlei Chen, Saining Xie, Yanghao Li, Piotr Doll{\'a}r, and Ross
  Girshick.
\newblock Masked autoencoders are scalable vision learners.
\newblock In {\em Proceedings of the IEEE/CVF Conference on Computer Vision and
  Pattern Recognition (CVPR)}, 2022.

\bibitem{he2016deep}
Kaiming He, Xiangyu Zhang, Shaoqing Ren, and Jian Sun.
\newblock Deep residual learning for image recognition.
\newblock In {\em Proceedings of the IEEE/CVF Conference on Computer Vision and
  Pattern Recognition (CVPR)}, 2016.

\bibitem{he2021transreid}
Shuting He, Hao Luo, Pichao Wang, Fan Wang, Hao Li, and Wei Jiang.
\newblock Transreid: Transformer-based object re-identification.
\newblock In {\em Proceedings of the IEEE/CVF International Conference on
  Computer Vision (ICCV)}, 2021.

\bibitem{hendrycks2019benchmarking}
Dan Hendrycks and Thomas Dietterich.
\newblock Benchmarking neural network robustness to common corruptions and
  perturbations.
\newblock {\em arXiv preprint arXiv:1903.12261}, 2019.

\bibitem{hendrycks2021nae}
Dan Hendrycks, Kevin Zhao, Steven Basart, Jacob Steinhardt, and Dawn Song.
\newblock Natural adversarial examples.
\newblock {\em Proceedings of the IEEE/CVF Conference on Computer Vision and
  Pattern Recognition (CVPR)}, 2021.

\bibitem{hoffer2020augment}
Elad Hoffer, Tal Ben-Nun, Itay Hubara, Niv Giladi, Torsten Hoefler, and Daniel
  Soudry.
\newblock Augment your batch: Improving generalization through instance
  repetition.
\newblock In {\em Proceedings of the IEEE/CVF Conference on Computer Vision and
  Pattern Recognition (CVPR)}, 2020.

\bibitem{jiang2021transgan}
Yifan Jiang, Shiyu Chang, and Zhangyang Wang.
\newblock Transgan: Two pure transformers can make one strong gan, and that can
  scale up.
\newblock {\em Advances in Neural Information Processing Systems (NeurIPS)},
  2021.

\bibitem{jung2020global}
Yunjae Jung, Donghyeon Cho, Sanghyun Woo, and In~So Kweon.
\newblock Global-and-local relative position embedding for unsupervised video
  summarization.
\newblock In {\em European Conference on Computer Vision (ECCV)}, 2020.

\bibitem{khan2022transformers}
Salman Khan, Muzammal Naseer, Munawar Hayat, Syed~Waqas Zamir, Fahad~Shahbaz
  Khan, and Mubarak Shah.
\newblock Transformers in vision: A survey.
\newblock {\em ACM computing surveys (CSUR)}, 54(10s):1--41, 2022.

\bibitem{kingma2015adam}
Diederik~P Kingma and Jimmy Ba.
\newblock Adam: A method for stochastic optimization.
\newblock In {\em International Conference on Learning Representations (ICLR)},
  2015.

\bibitem{kiyono2021shape}
Shun Kiyono, Sosuke Kobayashi, Jun Suzuki, and Kentaro Inui.
\newblock Shape: Shifted absolute position embedding for transformers.
\newblock {\em arXiv preprint arXiv:2109.05644}, 2021.

\bibitem{liang2021swinir}
Jingyun Liang, Jiezhang Cao, Guolei Sun, Kai Zhang, Luc Van~Gool, and Radu
  Timofte.
\newblock Swinir: Image restoration using swin transformer.
\newblock In {\em Proceedings of the IEEE/CVF International Conference on
  Computer Vision (ICCV)}, 2021.

\bibitem{liu2019roberta}
Yinhan Liu, Myle Ott, Naman Goyal, Jingfei Du, Mandar Joshi, Danqi Chen, Omer
  Levy, Mike Lewis, Luke Zettlemoyer, and Veselin Stoyanov.
\newblock Roberta: A robustly optimized bert pretraining approach.
\newblock {\em arXiv preprint arXiv:1907.11692}, 2019.

\bibitem{liu2021efficient}
Yahui Liu, Enver Sangineto, Wei Bi, Nicu Sebe, Bruno Lepri, and Marco Nadai.
\newblock Efficient training of visual transformers with small datasets.
\newblock {\em Advances in Neural Information Processing Systems (NeurIPS)},
  2021.

\bibitem{liu2022petr}
Yingfei Liu, Tiancai Wang, Xiangyu Zhang, and Jian Sun.
\newblock Petr: Position embedding transformation for multi-view 3d object
  detection.
\newblock {\em arXiv preprint arXiv:2203.05625}, 2022.

\bibitem{liu2021swin}
Ze Liu, Yutong Lin, Yue Cao, Han Hu, Yixuan Wei, Zheng Zhang, Stephen Lin, and
  Baining Guo.
\newblock Swin transformer: Hierarchical vision transformer using shifted
  windows.
\newblock In {\em Proceedings of the IEEE/CVF International Conference on
  Computer Vision (ICCV)}, 2021.

\bibitem{lu2022april}
Jiahao Lu, Xi~Sheryl Zhang, Tianli Zhao, Xiangyu He, and Jian Cheng.
\newblock April: Finding the achilles' heel on privacy for vision transformers.
\newblock In {\em Proceedings of the IEEE/CVF Conference on Computer Vision and
  Pattern Recognition}, pages 10051--10060, 2022.

\bibitem{mcinnes2018umap}
Leland McInnes, John Healy, and James Melville.
\newblock Umap: Uniform manifold approximation and projection for dimension
  reduction.
\newblock {\em arXiv preprint arXiv:1802.03426}, 2018.

\bibitem{rangrej2022consistency}
Samrudhdhi~B Rangrej, Chetan~L Srinidhi, and James~J Clark.
\newblock Consistency driven sequential transformers attention model for
  partially observable scenes.
\newblock In {\em Proceedings of the IEEE/CVF Conference on Computer Vision and
  Pattern Recognition}, pages 2518--2527, 2022.

\bibitem{russakovsky2015imagenet}
Olga Russakovsky, Jia Deng, Hao Su, Jonathan Krause, Sanjeev Satheesh, Sean Ma,
  Zhiheng Huang, Andrej Karpathy, Aditya Khosla, Michael Bernstein, et~al.
\newblock Imagenet large scale visual recognition challenge.
\newblock {\em International Journal of Computer Vision}, 115(3):211--252,
  2015.

\bibitem{shaw2018self}
Peter Shaw, Jakob Uszkoreit, and Ashish Vaswani.
\newblock Self-attention with relative position representations.
\newblock In {\em Proceedings of the Conference of the North American Chapter
  of the Association for Computational Linguistics (NAACL)}, 2018.

\bibitem{sohn2020fixmatch}
Kihyuk Sohn, David Berthelot, Nicholas Carlini, Zizhao Zhang, Han Zhang,
  Colin~A Raffel, Ekin~Dogus Cubuk, Alexey Kurakin, and Chun-Liang Li.
\newblock Fixmatch: Simplifying semi-supervised learning with consistency and
  confidence.
\newblock {\em Advances in neural information processing systems}, 33:596--608,
  2020.

\bibitem{su2021roformer}
Jianlin Su, Yu Lu, Shengfeng Pan, Bo Wen, and Yunfeng Liu.
\newblock Roformer: Enhanced transformer with rotary position embedding.
\newblock {\em arXiv preprint arXiv:2104.09864}, 2021.

\bibitem{tolstikhin2021mlp}
Ilya~O Tolstikhin, Neil Houlsby, Alexander Kolesnikov, Lucas Beyer, Xiaohua
  Zhai, Thomas Unterthiner, Jessica Yung, Andreas Steiner, Daniel Keysers,
  Jakob Uszkoreit, et~al.
\newblock Mlp-mixer: An all-mlp architecture for vision.
\newblock {\em Advances in Neural Information Processing Systems (NeurIPS)},
  2021.

\bibitem{touvron2021training}
Hugo Touvron, Matthieu Cord, Matthijs Douze, Francisco Massa, Alexandre
  Sablayrolles, and Herve Jegou.
\newblock Training data-efficient image transformers \& distillation through
  attention.
\newblock In {\em International Conference on Machine Learning (ICML)}, 2021.

\bibitem{vaswani2017attention}
Ashish Vaswani, Noam Shazeer, Niki Parmar, Jakob Uszkoreit, Llion Jones,
  Aidan~N Gomez, {\L}ukasz Kaiser, and Illia Polosukhin.
\newblock Attention is all you need.
\newblock In {\em Advances in Neural Information Processing Systems (NeurIPS)},
  2017.

\bibitem{wang2021pyramid}
Wenhai Wang, Enze Xie, Xiang Li, Deng-Ping Fan, Kaitao Song, Ding Liang, Tong
  Lu, Ping Luo, and Ling Shao.
\newblock Pyramid vision transformer: A versatile backbone for dense prediction
  without convolutions.
\newblock In {\em Proceedings of the IEEE/CVF International Conference on
  Computer Vision}, pages 568--578, 2021.

\bibitem{wang2022pvt}
Wenhai Wang, Enze Xie, Xiang Li, Deng-Ping Fan, Kaitao Song, Ding Liang, Tong
  Lu, Ping Luo, and Ling Shao.
\newblock Pvt v2: Improved baselines with pyramid vision transformer.
\newblock {\em Computational Visual Media}, 8(3):415--424, 2022.

\bibitem{wang2022tokencut}
Yangtao Wang, Xi Shen, Shell~Xu Hu, Yuan Yuan, James~L. Crowley, and Dominique
  Vaufreydaz.
\newblock Self-supervised transformers for unsupervised object discovery using
  normalized cut.
\newblock In {\em Proceedings of the IEEE/CVF Conference on Computer Vision and
  Pattern Recognition (CVPR)}, 2022.

\bibitem{wang2021end}
Yuqing Wang, Zhaoliang Xu, Xinlong Wang, Chunhua Shen, Baoshan Cheng, Hao Shen,
  and Huaxia Xia.
\newblock End-to-end video instance segmentation with transformers.
\newblock In {\em Proceedings of the IEEE/CVF Conference on Computer Vision and
  Pattern Recognition (CVPR)}, 2021.

\bibitem{wang2020position}
Yu-An Wang and Yun-Nung Chen.
\newblock What do position embeddings learn? an empirical study of pre-trained
  language model positional encoding.
\newblock In {\em Proceedings of the Conference on Empirical Methods in Natural
  Language Processing (EMNLP)}, 2020.

\bibitem{xie2020unsupervised}
Qizhe Xie, Zihang Dai, Eduard Hovy, Thang Luong, and Quoc Le.
\newblock Unsupervised data augmentation for consistency training.
\newblock {\em Advances in neural information processing systems},
  33:6256--6268, 2020.

\bibitem{xu2021vitae}
Yufei Xu, Qiming Zhang, Jing Zhang, and Dacheng Tao.
\newblock Vitae: Vision transformer advanced by exploring intrinsic inductive
  bias.
\newblock {\em Advances in Neural Information Processing Systems},
  34:28522--28535, 2021.

\bibitem{yang2021transformer}
Guanglei Yang, Hao Tang, Mingli Ding, Nicu Sebe, and Elisa Ricci.
\newblock Transformer-based attention networks for continuous pixel-wise
  prediction.
\newblock In {\em Proceedings of the IEEE/CVF International Conference on
  Computer Vision (ICCV)}, 2021.

\bibitem{ye2019cross}
Linwei Ye, Mrigank Rochan, Zhi Liu, and Yang Wang.
\newblock Cross-modal self-attention network for referring image segmentation.
\newblock In {\em Proceedings of the IEEE/CVF conference on computer vision and
  pattern recognition}, pages 10502--10511, 2019.

\bibitem{yin2021see}
Hongxu Yin, Arun Mallya, Arash Vahdat, Jose~M Alvarez, Jan Kautz, and Pavlo
  Molchanov.
\newblock See through gradients: Image batch recovery via gradinversion.
\newblock In {\em Proceedings of the IEEE/CVF Conference on Computer Vision and
  Pattern Recognition (CVPR)}, 2021.

\bibitem{yun2019cutmix}
Sangdoo Yun, Dongyoon Han, Seong~Joon Oh, Sanghyuk Chun, Junsuk Choe, and
  Youngjoon Yoo.
\newblock Cutmix: Regularization strategy to train strong classifiers with
  localizable features.
\newblock In {\em Proceedings of the IEEE/CVF International Conference on
  Computer Vision (ICCV)}, 2019.

\bibitem{zhang2018mixup}
Hongyi Zhang, Moustapha Cisse, Yann~N Dauphin, and David Lopez-Paz.
\newblock mixup: Beyond empirical risk minimization.
\newblock In {\em International Conference on Learning Representations (ICLR)},
  2018.

\bibitem{zhang2018unreasonable}
Richard Zhang, Phillip Isola, Alexei~A Efros, Eli Shechtman, and Oliver Wang.
\newblock The unreasonable effectiveness of deep features as a perceptual
  metric.
\newblock In {\em Proceedings of the IEEE/CVF Conference on Computer Vision and
  Pattern Recognition (CVPR)}, 2018.

\bibitem{zheng2021rethinking}
Sixiao Zheng, Jiachen Lu, Hengshuang Zhao, Xiatian Zhu, Zekun Luo, Yabiao Wang,
  Yanwei Fu, Jianfeng Feng, Tao Xiang, Philip~HS Torr, et~al.
\newblock Rethinking semantic segmentation from a sequence-to-sequence
  perspective with transformers.
\newblock In {\em Proceedings of the IEEE/CVF Conference on Computer Vision and
  Pattern Recognition (CVPR)}, 2021.

\bibitem{zhu2020r}
Junyi Zhu and Matthew Blaschko.
\newblock R-gap: Recursive gradient attack on privacy.
\newblock In {\em International Conference on Learning Representations (ICLR)},
  2021.

\bibitem{zhu2021deformable}
Xizhou Zhu, Weijie Su, Lewei Lu, Bin Li, Xiaogang Wang, and Jifeng Dai.
\newblock Deformable detr: Deformable transformers for end-to-end object
  detection.
\newblock In {\em International Conference on Learning Representations (ICLR)},
  2021.

\end{thebibliography}
